%% file: main.tex
\documentclass{article}
\usepackage{amssymb}
\usepackage{cite}
\usepackage{amsmath,amssymb,amsfonts}
\usepackage{algorithmic}
\usepackage{graphicx}
\usepackage{textcomp}
\usepackage{xcolor}
\usepackage{xspace}
\usepackage{siunitx}
\usepackage{wrapfig}
\usepackage{booktabs}
\usepackage{multirow}
\usepackage{enumitem}
\usepackage{verbatim}
\usepackage{makecell}
\usepackage[font=footnotesize, skip=7pt]{caption}

\newcommand{\bA}{\mathbf{A}}
\newcommand{\bC}{\mathbf{C}}
\newcommand{\ba}{\mathbf{a}}
\newcommand{\bd}{\mathbf{d}}

\newcommand{\bo}{\mathbf{o}}
\newcommand{\bp}{\mathbf{p}}

\newcommand{\bx}{\mathbf{x}}
\newcommand{\bI}{\mathbf{I}}

\newcommand{\method}{DiffusiveGRAIN\xspace}

\usepackage[preprint]{corl_2025} 

\title{Granular Loco-Manipulation: Repositioning Rocks Through Strategic Sand Avalanche}

\author{
  \textbf{Haodi Hu}$^1$,
  \textbf{Yue Wu}$^1$,
  \textbf{Feifei Qian}$^{1,}$$^\dagger$,
  \textbf{Daniel Seita}$^{1,}$$^\dagger$ \\
  feifeiqi@usc.edu\\
  $\dagger$ Equal Advising\\
  $^1$University of Southern California, United States
}

\begin{document}
\maketitle


\vspace{-15pt}
\begin{abstract}
    Legged robots have the potential to leverage obstacles to climb steep sand slopes. However, efficiently repositioning these obstacles to desired locations is challenging. Here we present \method, a learning-based method that enables a multi-legged robot to strategically induce localized sand avalanches during locomotion and indirectly manipulate obstacles. 
    We conducted 375 trials, systematically varying obstacle spacing, robot orientation, and leg actions in 75 of them. Results show that movement of closely-spaced obstacles exhibit significant interference, requiring joint modeling. In addition, different multi-leg excavation actions could cause distinct robot state changes, necessitating integrated planning of manipulation and locomotion.
    To address these challenges, \method includes a diffusion-based environment predictor to capture multi-obstacle movements under granular flow interferences and a robot state predictor to estimate changes in robot state from multi-leg action patterns. Deployment experiments (90 trials) demonstrate that by integrating the environment and robot state predictors, the robot can autonomously plan its movements based on loco-manipulation goals, successfully shifting closely located rocks to desired locations in over 65\% of trials. Our study showcases the potential for a locomoting robot to strategically manipulate obstacles to achieve improved mobility on challenging terrains. Supplementary material is available at \url{https://sites.google.com/view/diffusivegrain/home}.
\end{abstract}

\keywords{Granular media, avalanche dynamics, diffusion models, legged robots}  


\section{Introduction}

\begin{figure}[hbtp]
    \centering
    \includegraphics[width=0.90\textwidth]{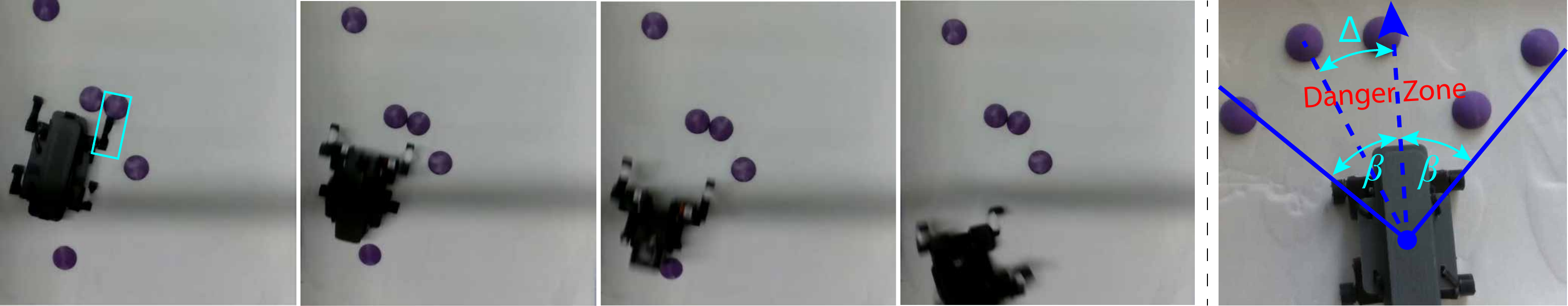}
    \caption{
        Left (4 images): An example of the robot flipping backward (over its two back legs) on a steep sand slope due to undesired leg-obstacle contact. The cyan rectangle highlights the leg's contact with the obstacle. Right: Stepping on undesired  locations can result in a high risk of robot slipping, stuck, or flipping backwards. 
    }
    \label{fig:robot failed demo}
    \vspace{-12pt}
\end{figure}

Natural environments contain deformable sand, steep inclines, and large rocks and boulders, which present significant challenges for terrestrial robot locomotion. 
Recent robotics research has explored bio-inspired~\cite{schiebel2017collisional,schiebel2019mechanical} ``obstacle-aided locomotion'' strategies, enabling robots to utilize interactions and collisions with large rocks and boulders to improve mobility on complex terrains~\cite{qian2019obstacle,othayoth2020energy,rieser2019,wang2023mechanical,haodi-obstacle,hu2024multirobot}.  
While obstacle-aided locomotion offers promising opportunities for robots~\cite{ramesh2020modulation,chakraborty2022planning,haodi-obstacle} to negotiate challenging terrains, these strategies often rely on specific leg-obstacle contact positions~\cite{qian2015anticipatory,rieser2019}. Mischosen contact positions can lead to catastrophic failures due to slipping, getting stuck, or even flipping over (Fig.~\ref{fig:robot failed demo}). As a result, their effectiveness depends heavily on the availability and spatial distribution of rocks and boulders, which can vary unpredictably across natural terrains. 

A recent study~\cite{hulearning} found that by strategically triggering localized sand avalanches~\cite{barker2000two, pudasaini2007avalanche, gravish2014effect}, a robotic leg can generate granular flow on steep sand slopes, repositioning large rocks and boulders to desired locations. This introduces a new approach for indirect obstacle manipulation through granular flow. 
By integrating this capability with obstacle-aided locomotion strategies, legged robots can modify their environments to enhance traversability. However, most existing granular manipulation methods focus either on homogeneous granular media~\cite{pavlov2019soil,wang2023dynamicresolution,xue2023neuralfield,schenck2017learning} or assume that obstacle movements under granular flow are independent of robot locomotion and movement of other obstacles~\cite{hulearning}. How to extend granular flow based manipulation capabilities to locomoting robots and substrates with densely-distributed obstacles remains unexplored. 

\begin{figure*}[t]
    \centering
    \includegraphics[width=1.00\textwidth]{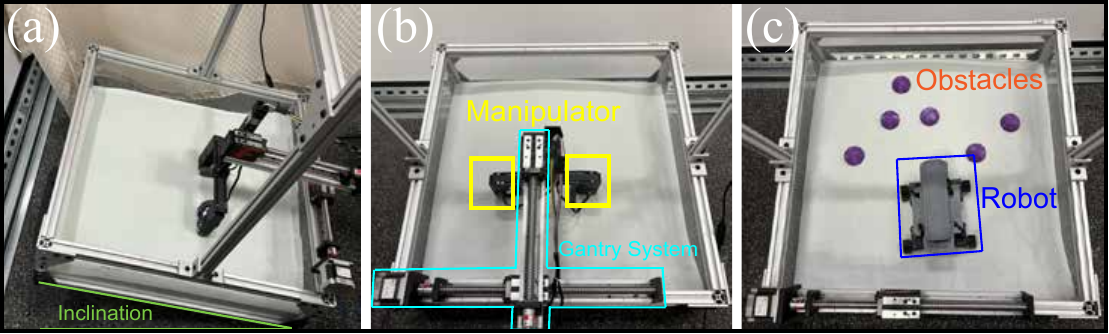}
    \caption{Experiment environment, with (a) a side view of the granular trackway with an inclination angle of $\Phi=20$ degrees; (b) the granular trackway with two robotic legs mounted on an actuated gantry system; (c) the robot (\emph{not} the manipulator in (b)) in the granular trackway, and 3D-printed obstacles (purple semi-spheres). 
    }
    \label{fig:experiment environment}
    \vspace{-14pt}
\end{figure*}

To address this, we investigate multi-obstacle manipulation on granular slopes using a locomoting quadrupedal robot, where each leg functions as an excavator to interact with sand and indirectly reposition obstacles (Fig. \ref{fig:experiment environment}C). Our findings reveal two critical challenges: (1) obstacle movement can be influenced by nearby obstacles on the sand, requiring a new method to resolve multiple obstacle movement under leg excavation simultaneously; and (2) the robot state (i.e., its position and orientation) can also be substantially affected by multi-leg excavation actions, necessitating a method that predicts and plans for both environment and robot state changes.

To tackle these challenges, we propose \method, a learning-based method that enables multi-legged robots to reposition densely-distributed rocks on sand slopes to desired locations during its locomotion. \method has a novel granular media dynamics predictor that learns both obstacle and robot movement under leg excavation actions. 
It takes as input, a depth image of the environment state, and a separate image representation of the leg excavation action. 
The predictor has two U-Net~\cite{ronneberger2015unet} models. One is a diffusion~\cite{sohldickstein2015deepunsupervisedlearningusing,yang2023diffusion,ho2020denoisingdiffusionprobabilisticmodels} model that predicts the environment state change, and the other predicts the robot state change. 
During deployment, the dynamics predictor predicts the robot state change given its action. Since we train the environment predictor using data from a ``manipulator'' (see Fig.~\ref{fig:experiment environment}B) instead of the robot (Fig.~\ref{fig:experiment environment}C), we adjust the action representation passed to the environment predictor, which then predicts granular surface depth change. 

In summary, our contributions are:
(i) A novel manipulation predictor using a U-Net with diffusion for a legged robot to predict avalanche dynamics of a sand slope with obstacles via leg-terrain interaction. 
(ii) A novel locomotion predictor based on a U-Net for a legged robot to predict its state change on a sand slope given the robot's action.
(iii) Integration of the proposed manipulation and locomotion predictors to enable a legged robot to achieve loco-manipulation tasks on a sand slope.

\section{Preliminaries}\label{sec:preliminaries}
\textbf{Experiment Setup:}
Fig.~\ref{fig:experiment environment} illustrates our physical gantry system.  
The granular trackway is (\SI{60}{\centi\meter} $L$ $\times$ \SI{60}{\centi\meter} $W$ $\times$ \SI{20}{\centi\meter} $D$) and contains model granular medium (Grainger, \SI{0.3}{\milli\meter}  glass beads). The \SI{0.3}{\milli\meter} particle size is similar to those observed in natural deserts, and behave qualitatively similar to natural sand and are widely used in granular robotics studies~\cite{maladen2009undulatory,li2013terradynamics,finn2016particle}.
The granular trackway can be tilted up to 35 degrees to emulate a wide variety of sand slopes in natural environments~\cite{gravish2014effect}.
In this experiment, we chose to use 
a slope angle $\Phi = 20$ degrees, which is close to the angle of repose~\cite{albert1997maximum} of our granular material and facilitates the study of avalanche dynamics. 
We mount a top-down RealSense 435i RGBD camera to record granular flow and obstacle movement.

\textbf{Manipulator versus Robot:} 
To study avalanche dynamics and object movement upon different leg excavation actions, we build a gantry system with two linear actuators, which move along the $x$ and $y$ axes. This gantry moves a \emph{manipulator} on a 2D surface \emph{above} the granular slope. 
This manipulator contains two C-shape robot leg motors with rotation centers \SI{1.0}{\centi\meter} above the granular surface. Each has a diameter of \SI{6.0}{\centi\meter} and a width of \SI{1.5}{\centi\meter}. 
In this paper, the ``manipulator'' is distinct from the actual \emph{robots} (i.e., quadrupeds) we use in experiments.
The manipulator's two C-shape leg designs match the front two C-shape legs of the robot, and it can also execute the same excavation action as the robot. 
The manipulator enables faster and safer data collection compared to using the robot. 

\textbf{Action Space:}
The action space $\mathcal{A}$ for the robots (and manipulators) has six actions, each containing a different group of the four robot legs, (Fig.~\ref{fig:pre-exp} bottom left): (1) \emph{Left Front Excavation (LFE)}, where the left front (LF) leg rotates backwards relative to the robot heading; (2) \emph{Right Front Excavation (RFE)}\, where the right front (RF) leg rotates backwards through the granular medium; (3) \emph{Left Pair (LP)}, where the robot's left front (LF) and left hind (LH) legs rotate backwards synchronously; (4) \emph{Right Pair (RP)}, where the robot's right front (RF) and right hind (RH) legs rotate backwards synchronously; (5) \emph{Front Pair (FP)}, where the robot's left front (LF) and right front (RF) legs rotate backwards synchronously; and (6) \emph{All Four (AF)}, where all four robot legs rotate backwards synchronously. For all actions, the rotating angular speed of each leg is constant, $1 \pi~rad/s$. 
\vspace{-5pt}
\section{Relevant Work and Knowledge Gap}
\label{sec:challenges_grain}
\vspace{-5pt}

Prior work has investigated terrain manipulation techniques, wherein a robotic leg or wheel supports body weight and actively reconfigures, compacts, or fluidizes the substrate to bolster stability and traction~\cite{pavlov2019soil,shrivastava2020material, karsai2022real, kerimoglu2024learning,hulearning}. Such innovations leverage terra-dynamics~\cite{li2013terradynamics}, the study of movement in granular substrates to inform real-time control decisions, enabling robots to traverse and reshape their environment for more efficient locomotion~\cite{schwarz2016supervised}. This approach has broad implications for challenging real-world operations, from extraterrestrial exploration to search-and-rescue missions. 
One particularly interesting work in this domain is GRAIN~\cite{hulearning}, a learning-based approach that predicts sand slope change based on leg excavations. 
GRAIN used a Vision Transformer (ViT)~\cite{dosovitskiy2020image} to process image representations of granular dynamics and robot excavation actions. 
The ViT predicts 2D object movement on a granular medium slope corresponding to different excavation actions.

While GRAIN worked well for single leg, single obstacle manipulation, it assumes the following which may be invalid for multi-leg, multi-obstacle scenarios:
(i) GRAIN assumed that obstacle movements are independent and not affected by the movement or granular flow of adjacent obstacles. 
(ii) GRAIN assumed that obstacle movement for each leg excavation could be trained independently and did not consider the effects of simultaneous multi-leg manipulation.
(iii) GRAIN assumed that the robot state remained unaffected under leg excavation actions, which may be invalid since a locomoting robot could slip and turn as it performs excavations. In addition, such state change may depend on which leg (or legs) was used for excavation. 
To test these assumptions under multi-obstacle, multi-leg manipulation tasks, we conducted two sets of experiments for GRAIN. 

\begin{figure*}[t]
    \centering
    \includegraphics[width=0.85\linewidth]{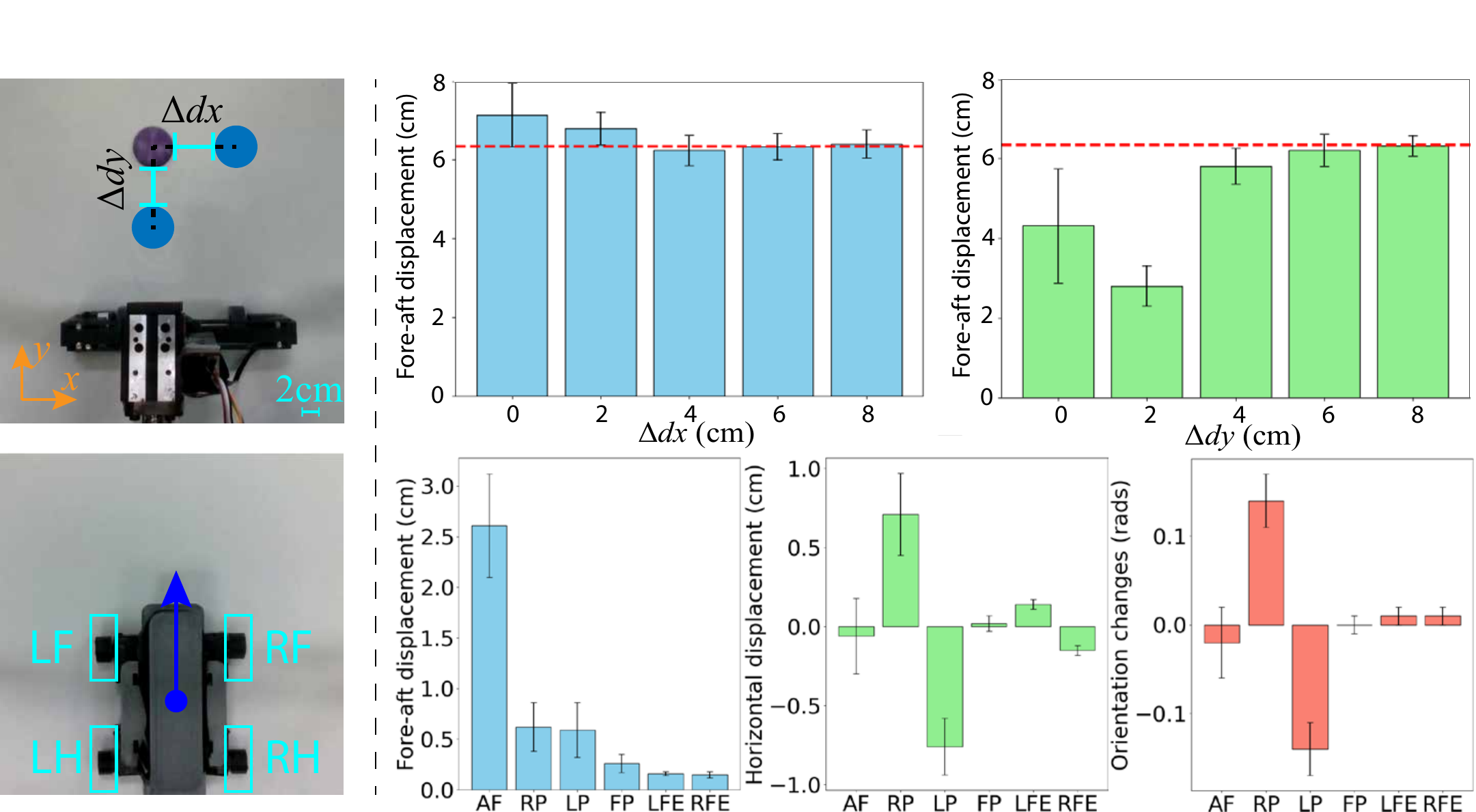}
    \caption{Experiment setup for investigating GRAIN~\cite{hulearning} (see Sec.~\ref{sec:challenges_grain}). 
    Left top: multi-obstacle manipulation experiment setup, where $\Delta dx$ is the lateral distance and $\Delta dy$ is the fore-aft distance between the two obstacles; Left bottom: the robot state change experiment, where we investigate the robot locomotion state change under 6 different combinations of leg excavation actions; Right top: 2 obstacles distance in horizontal and fore-aft directions affects obstacle movement, red dash line represents the obstacle movement without the affect of the other obstacle; Right bottom: statistics of robot state change by different robot actions.}
    \label{fig:pre-exp}
    \vspace{-12pt}
\end{figure*}

\textbf{Multi-obstacle with systematically-varied distance}, where we systematically varied the positions between two adjacent obstacles from 0 to 8cm, with a 2cm increment (Fig. \ref{fig:pre-exp} Left).
Experimentally measured obstacle movement suggested that as the distance between obstacles decreases, the influence on obstacle movement becomes stronger (Fig.~\ref{fig:pre-exp} top row right, 2cm and 0cm). This effect is particularly significant when additional obstacles are added in the direction of sand inclination (Fig. \ref{fig:pre-exp} top row left, $y$ axis), where the displacement of an obstacle was observed to decrease to 67$\%$ and 42$\%$ as compared to no obstacle (Fig.~\ref{fig:pre-exp} top row right, 2cm and 0cm, respectively). 
This suggests that treating obstacle movements as independent~\cite{hulearning} only works when obstacles are sufficiently far apart.
When obstacles are close, adjacent ones significantly affect avalanche propagation~\cite{borzsonyi2008avalanche}. 

\textbf{Robot state change under leg excavation actions}, where we tested six sets of multi-leg excavation actions (see Sec.~\ref{sec:preliminaries}) to investigate the effect of leg manipulation actions on robot locomotion. Experimental measurements from the 6 multi-leg manipulation actions showed that the \emph{AF} action resulted in the largest displacement in the robot fore-aft position. The \emph{RP} and \emph{LP} actions resulted in the largest change in the robot orientation and lateral position, while not changing much of the robot fore-aft position. The \emph{FP}, \emph{LFE}, \emph{RFE} are manipulation actions that trigger avalanche behavior at different locations while keeping the robot position change small. 
See Fig.~\ref{fig:pre-exp} (bottom right) for aggregated results over 3 trials for each robot action, and Fig.~\ref{fig:comparison of GRAIN and current on locomotion} in the Appendix for visualizations. These results indicated that the change in robot state can sensitively depend on the group of robot legs used in manipulation. For loco-manipulation tasks, it is critical to consider and plan these robot state changes jointly with manipulation goals. 

These experiments demonstrated the limitations of GRAIN, and highlighted two key challenges that require new methods: (i) the need to consider the granular flow influence from adjacent obstacles and (ii) the need to jointly plan manipulation and locomotion actions. 


\vspace{-6pt}
\section{\method: Learning to Predict Obstacle Movement} 
\label{Sec: method}

\begin{figure*}[htbp]
    \centering
    \includegraphics[width=1.00\textwidth]{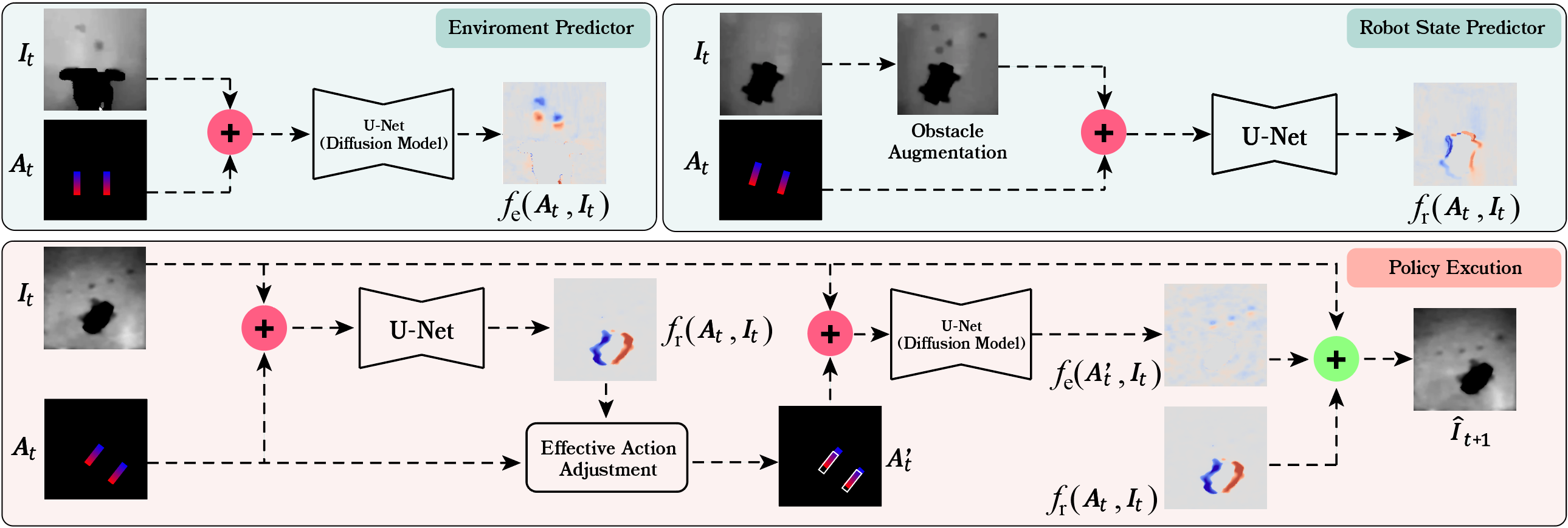}
    \caption{System overview. The \emph{environment predictor} $f_e$ uses a diffusion model (with a U-Net backbone) to predict the depth change of the environment given the depth image and action. The \emph{robot state predictor} $f_r$ uses a U-Net to predict the robot state change given the robot state and action. During \emph{policy execution}, given the predicted robot state change, we introduce an ``Effective Action Adjustment (EAA)" (see Sec.~\ref{ssec:EAA}). 
    We then combine the updated robot action image with the depth image to the trained diffusion model and get the predicted depth image change. We combine this with the predicted robot state and the original depth image to get the predicted next depth image. The red addition symbols represent channel-wise image concatenation operation, and the green addition symbols represent the image combination method as described in Sec.~\ref{ssec:policy_execution}. 
    }
    \label{fig:overview}
    \vspace{-10pt}
\end{figure*}

\subsection{State Representation and Robot Action Space}


A top-down camera provides a depth image $\bI_t$ at time $t$ (see Fig.~\ref{fig:overview} for examples). We denote the \emph{robot state} in $\bI_t$ as $\bx_t \in \mathbb{R}^3$, consisting of 2D position and 1D orientation. We denote the starting state as $\bx = (x^*, y^*, \phi^*)$. 
The \emph{obstacles' states} in $\bI_t$ is represented as $\bo_t$ where $\bo_t^i \in \mathbb{R}^2$ represents the $i$-th obstacle state (represented as a 2D position). The robot action space $\mathcal{A}$ has 6 actions as described in Sec.~\ref{sec:preliminaries}. 
To test \method, we propose to study the following tasks:

\emph{Manipulation}: A robot starting at $\bx$ needs to manipulate $N$ obstacles with initial states $\bo^i = (o^i_x, o^i_y)$ to target states $\bd^o_i = (d^i_x, d^i_y)$, where $i$ is an integer, i.e., $i \in \{1, N\}$. 

\emph{Locomotion}: A robot starting at $\bx$ needs to travel to a desired location $\bd^r = (d_x, d_y)$.

\emph{Loco-manipulation}: A robot at $\bx$ needs to travel to a desired location $\bd^r = (d_x, d_y)$ and manipulate $N$ obstacles with initial states $\bo^i = (o^i_x, o^i_y)$ to targets $\bd^o_i = (d^i_x, d^i_y)$, where $i \in \{1, N\}$. 
We consider two variants: \emph{in-distribution} and \emph{out-of-distribution}. The former uses sand slope angles and objects from training, while the latter tests generalization by using different angles and objects.

\subsection{Environment State and Robot State Predictors}
\label{ssec:state_predictors}

\textbf{Environment State Predictor $f_e$.}
We improve over GRAIN~\cite{hulearning} by using a diffusion model instead of a Vision Transformer~\cite{dosovitskiy2020image} and by directly collecting data with multiple obstacles on the granular slope. 
Our $f_e$ has a U-Net backbone~\cite{ronneberger2015unet} diffusion model. The inputs are the depth image $\bI_t$ and an RGB image $\bA_t$ representing the robot action. 
For $\bA_t$, we use a space-aligned gradient color region from blue to red to represent the robot leg interaction area with the granular slope. The color region's length is the effective action length (\SI{12.0}{\centi\meter}) and its width is the robot leg width (\SI{1.5}{\centi\meter}). 
We train $f_e$ to predict the change in the depth image of the granular slope surface $f_e(\bI_t, \bA_t)$. The predicted image is converted to a grayscale image, which we then add to the original input depth image to get the predicted depth image of the environment for the \emph{next} state: $\bI_t + f_e(\bI_t, \bA_t)$.

\textbf{Robot State Predictor $f_r$.}
We use a second U-Net, $f_r$, to predict the robot state change given its action and the environment state. 
As in $f_e$, the inputs are the depth image $\bI_t$ and the RGB image action representation $\bA_t$.  
During training, we use OpenCV code~\cite{opencv_library} to augment the input data to $f_r$ by adding obstacles to the depth image of the collected dataset while keeping the same label. As a result, the U-Net learns from the input depth images with extra obstacles. We use this method based on our observation during the experiments that the obstacles do not noticeably affect robot's state change unless the robot leg directly contacts obstacles. The output is the predicted robot state change $f_r(\bI_t, \bA_t)$. We can similarly convert this to a grayscale and obtain the predicted depth image representing the robot, $\bI_t + f_r(\bI_t, \bA_t)$, for the \emph{next} state. See Fig.~\ref{fig:overview} for an overview.


\subsection{Effective Action Adjustment (EAA)}
\label{ssec:EAA}

We observe that \emph{robot} action-triggered sand avalanche behavior can significantly differ from the sand avalanche behavior triggered by the \emph{manipulator} with the same excavation action, especially when the robot executes \emph{AF, LP, RP} actions. The reason is that robot leg excavation actions can lead to a different amount of advancement or slippage in each leg, resulting in significant changes to the robot state during excavation. 
We propose an Effective Action Adjustment (EAA) method to compensate for this prediction error. Based on the robot action, we know there are two leg-sand interaction events in a full rotation of a robot leg, and the robot state change is because the robot leg rotation provides a robot propulsion and rotation force to change its position and orientation. The robot has an initial state $\bx_0$ and during the first leg-sand interaction the robot state changes to $\bx_1$ and later changes to $\bx_2$ during the second leg-sand interaction. The EAA assumes the leg excavation action triggered sand avalanche during the robot transition from $\bx_0$ to $\bx_2$ is equal to the leg excavation action triggered a sand avalanche at the fixed robot state $\bx_1$. To get $\bx_1$ in the \method policy execution stage, we assume $\bx_1 = \frac{\bx_0 + \bx_2}{2}$. As a result, during policy execution, we update the action image $A_t$ by changing the shaded color area corresponding to the robot state $\bx_1$, where the $\bx_0$ is extracted from $\bI_t$ and the $\bx_2$ is extracted from $f_r(\bA_t, \bI_t)$. See the Appendix~\ref{app:baseline} for details. 

\subsection{Cost Functions for Different Modes}\label{sec:costfun}

\textbf{Cost function for locomotion:} During robot locomotion mode, we consider (i) the distance between the robot CoM and target location and (ii) robot safety. For the distance, given the target location and robot position on the sand, $\bd^r$ and $\bx_p = (x^*, y^*)$, we define the cost as:
\begin{equation}
    \bC_{rt} = \| \bx_p - \bd^r \|_2.
    \label{eq:cost for distance}
\end{equation}
For robot safety, we consider (i) the distance of the obstacles and robot CoM and (ii) the relative angle between the line connecting the obstacle and robot heading direction. 
Given the robot state $\bx = (x^*, y^*, \phi^*)$ and an obstacle location $\bo = (o_x, o_y)$, the angle difference between the line connecting the obstacle and robot heading direction, $\Delta$, is defined as $\Delta \;=\; \phi - \phi^*$, where $\phi \;=\; \mathrm{atan2}\bigl(o_y - y^*,\, o_x - x^*\bigr)$. 
We use these quantities because according to a previous study, leg-obstacle contact position can significantly influence robot locomotion outcomes~\cite{qian2015dynamics}. While stepping on certain locations of obstacles could aid locomotion, the robot can also be at high risk of flipping over or getting high centered on the obstacle when misstepping on undesired obstacle locations. Based on qualitative observations of these failure events (Fig.~\ref{fig:robot failed demo}), we introduce a penalty for obstacles within the ``danger zone" of the robot heading direction. We define the robot safety cost $\bC_{rs}$ as:
\begin{equation}
\label{eq:cost for safety}
\begin{array}{l@{\hskip 1.0cm}l}
\bC_{rs} = \sum_i \dfrac{\mathrm{AngFactor}(\Delta_i)}{\| \bx_p - \bo^i \|_2} ,
&
\mathrm{AngFactor}(\Delta) =
\begin{cases}
\exp\bigl(\alpha(\beta - |\Delta|)\bigr), & \text{if } |\Delta| \le \beta, \\[2pt]
1, & \text{otherwise}.
\end{cases}
\end{array}
\end{equation}
See Fig.~\ref{fig:robot failed demo} for $\beta$ and $\Delta$ visualizations. We formulate the cost for locomotion by linearly combining Eq.~\ref{eq:cost for distance} and Eq.~\ref{eq:cost for safety}: $\bC_l = w_1  \bC_{rt} + w_2  \bC_{rs}$, where we use $w_1 = 0.6$, $w_2 = 0.4$, $\alpha = 4$, and $\beta = \frac{\pi}{4}$.

\textbf{Cost function for manipulation:} During robot manipulation mode, we consider (i) the distance between obstacle positions and targets, and (ii) robot safety. For the distance between obstacle positions and targets, given the target locations on the sand slope, $\bd^o_i$, the cost is defined as in Eq.~\ref{eq:cost for obstacle manipulation distance}: 
\begin{equation}
    \bC_o = \sum_i \| \bo^i - \bd^o_i \|_2.
    \label{eq:cost for obstacle manipulation distance}
\end{equation}
The robot safety cost is the same as in locomotion mode (Eq.~\ref{eq:cost for safety}). 
The manipulation mode cost, $\bC_m$, is a linear combination of Eq.~\ref{eq:cost for safety} and Eq.~\ref{eq:cost for obstacle manipulation distance}: $\bC_m = w_3  \bC_o + w_4  \bC_{rs}$; we use $w_3 = 0.8$ and $w_4 = 0.2$.

\subsection{Policy Planning and Execution}
\label{ssec:policy_execution}

Previous work~\cite{hulearning} demonstrated the potential of a greedy strategy for certain manipulation tasks. While this can be efficient in simpler scenarios, it often fails in more complex loco-manipulation tasks that require jointly considering locomotion and manipulation. Our preliminary experiments indicate that to effectively leverage sand avalanches—particularly by controlling their impact regions—the robot must plan multiple steps in advance. Specifically, multi-step action planning ensures that locomotion and manipulation are coordinated in ways that reposition obstacles advantageously on the sand slope, which is crucial for successful loco-manipulation. As a result, we adopt a receding-horizon planning method that minimizes the cumulative cost: $\sum^{t=3}_{t=0} \gamma^t \bC(\bx_t, \bo_t, \ba_t)$ by planning actions for the next 4 steps, where $\gamma$ is the discount rate and $\bC$ is the cost function for a given task (\textit{i.e.} manipulation or locomotion or both). We set $\gamma = 0.8$ in this paper.


\textbf{Policy Execution.} During deployment, given the depth image $\bI_t$ and robot action $\bA_t$, we first use $f_r$ (Sec.~\ref{ssec:state_predictors}) to predict the robot state change, $f_r(\bI_t, \bA_t)$. 
We then use our EAA (Sec.~\ref{ssec:EAA}) to update the robot action representation image to $\bA_t'$.  
To combine the predicted robot state change and the predicted environment state change to get the system's predicted next state (i.e., depth image) $\hat \bI_{t+1}$, we introduce another step to proceed with the image representation of environment state and robot state. Specifically, we identify the pixel set that represents the robot at time $t$ in $\bI_t$, $\bp_t$, and we use the same method to identify the pixel set that represents the robot in $\bI_t + f_r(\bI_t, \bA_t)$, $\bp_{t+1}$. We use the $\bp_{t+1}$ to replace the pixel in the depth image representation of the predicted environment observation image, $\bI_t + f_e(\bI_t, \bA_t^\prime)$, and we replace the pixel in $\bp_t$ but not in $\bp_{t+1}$ with the mean value of surrounding pixels in $\bI_t + f_e(\bI_t, \bA_t^\prime)$. With this additional step, we combine the diffusion model's predicted environment next state with the U-Net predicted robot next state and get the depth image that represents the system's next state $\hat \bI_{t+1}$. In this work, the robot only has six actions (see Sec.~\ref{sec:preliminaries}), so we iterate through all possible robot action combinations for the next 4 steps.


\section{Evaluation}
\label{sec:eval}

\subsection{Collecting Real-World Training Data}



We use the system shown in Fig.~\ref{fig:experiment environment} (see Sec.~\ref{sec:preliminaries} for background). 
To scalably collect training data while reducing the risk of interruptions if obstacles enter the robot's danger zone (see Fig.~\ref{fig:robot failed demo}), we separate data collection into (i) \emph{leg manipulation data} using the gantry and (ii) \emph{locomotion data} using the quadruped robot.
For \emph{leg manipulation}, we collect 240 ($60 \times 3 + 60$) trials, resulting in 24,590 images. This includes 60 trials for each gantry manipulator manipulating obstacles at orientations of 0, 15, and 30 degrees, with the left and right manipulators executing excavation actions. We also do 10 trials each for the gantry manipulator manipulating obstacles at orientations of 0, 15, and 30 degrees with just the left or right leg. 
For \emph{robot locomotion}, we detach the gantry system to free space for the robot. We collect 60 trials, resulting in a total of 13,480 images. We do 5 trials for each robot orientation of 0, 15, and 30 degrees with the \emph{AF} action, and 3 trials for each robot orientation of 0, 15, and 30 degrees with the other 5 actions. 

\subsection{Baselines and Evaluation Protocol}
\begin{figure*}[t]
    \centering
    \includegraphics[width=1.0\textwidth]{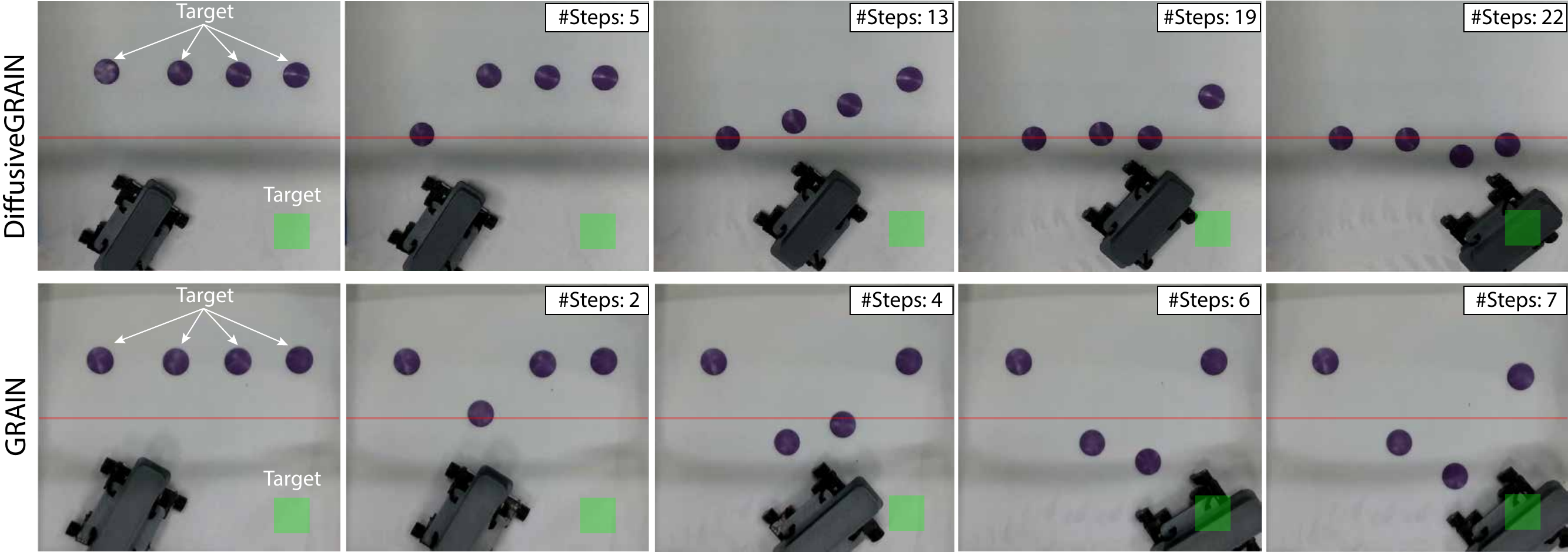}
    \caption{An example robot loco-manipulation trial for \method and GRAIN. The robot must bring 4 obstacles below the red horizontal line while also moving to the target marked with the green square. In \method, the robot achieved both locomotion and manipulation at step 22. In GRAIN, the robot achieved its locomotion task but only moved the middle 2 obstacles below the red line, and thus failed in manipulation.
    }
    \label{fig:single robot locomanipulation experiment}
    \vspace{-12pt}
\end{figure*}

\emph{Robot movement prediction Baseline:} For the locomotion model, we use GRAIN~\cite{hulearning} to predict the robot state change. GRAIN takes the depth image concatenated with the robot action representation image as input and outputs a $1\times 3$ matrix containing the robot's 2D position and 1D orientation.
\emph{Manipulation Planning Baseline:} 
We also adapt GRAIN~\cite{hulearning} to this setup. We trained GRAIN with the new dataset and output the coordinates of each obstacle. The GRAIN framework was originally designed to train on depth images with only one obstacle, resulting in a $1\times 2$ matrix corresponding to obstacle positions on the x-axis and y-axis. However, we have 1 to 5 obstacles. Consequently, we provided the ground truth number of obstacles to GRAIN and modified its output to be a $5\times 2$ matrix. When there are $N < 5$ obstacles, GRAIN only considers the first $N$ rows in the matrix.

We evaluate using Euclidean distance. To get the position of each obstacle and robot, we estimate their center of mass. For manipulation, we use the Euclidean distance among obstacle positions and their targets. For locomotion, we measure the distance between the robot and the locomotion target. 


\subsection{Evaluation Results}
\vspace{-6pt}


\begin{wraptable}{r}{0.55\textwidth}
    \vspace{-13pt}
    \centering
    \small
    \begin{tabular}{>{\centering\arraybackslash}m{1.5cm} >{\centering\arraybackslash}m{1.15cm} >{\centering\arraybackslash}m{1.75cm} >{\centering\arraybackslash}m{1.6cm}}
        \toprule
        \textbf{Task} & \textbf{Method} & \textbf{MAE} (cm) & \textbf{Success} (\%)\\
        \midrule
        Manipulation & \makecell{GRAIN \\ Ours} & \makecell{2.71($\pm$ 0.91) \\ 2.03($\pm$ 0.84)} & \makecell{60 \\ 80} \\
        \midrule
        Locomotion & \makecell{GRAIN \\ Ours} & \makecell{1.34($\pm$ 0.46) \\ 1.21($\pm$ 0.38)} & \makecell{80 \\ 90} \\
        \midrule
        Loco-manipulation & \makecell{GRAIN \\ Ours} & \makecell{3.24($\pm$ 1.24) \\ 2.31($\pm$ 0.98)} & \makecell{20 \\ 70} \\
        \bottomrule
    \end{tabular}
    \caption{
        We compare the performance of \method and GRAIN during policy execution on the tasks we study.
    }
    \label{tab:comparison of GRAIN and DiffusiveGRAIN on policy execution}
    \vspace{-9pt}
\end{wraptable}

\emph{Manipulation:} 
\method results in 8/10 success, while GRAIN has 6/10 success. 
See Tab.~\ref{tab:comparison of GRAIN and DiffusiveGRAIN on policy execution} for results. Fig.~\ref{fig:manipulation experiment} (in the Appendix) shows top-view images during policy execution for a robot manipulation trial with 4 obstacles (i.e., objects). In both trials, the robot must indirectly manipulate all 4 obstacles to bring them to a designated goal region near the robot. 
The improved performance suggested the importance for including training data with multiple obstacles, especially when obstacle distances are small (see Fig.~\ref{fig:statisitcs of GRAIN on obstacles with diff distance}, also in the Appendix). 

\emph{Locomotion:} 
\method obtains 9/10 success, while GRAIN gets 8/10 success. 
Fig.~\ref{fig:locomotion experiment} (in the Appendix) shows 1 locomotion trial per method. 

\emph{Loco-manipulation (in-distribution):} 
Using \method results in 7/10 success, compared to 2/10 for GRAIN. 
Fig.~\ref{fig:single robot locomanipulation experiment} shows 1 loco-manipulation trial using \method and GRAIN. The robot must move all 4 obstacles to a target region (below the red line) and navigate to a target location (marked by a green square). The result suggests that \method enables the robot to successfully plan simultaneous manipulation and locomotion, and highlights the importance of combining the environment state and robot state predictors for loco-manipulation tasks, as mischosen leg excavations may lead to undesired robot state change and preclude future manipulation options.

\begin{figure}[t]
  \centering
  \includegraphics[width=1.0\linewidth]{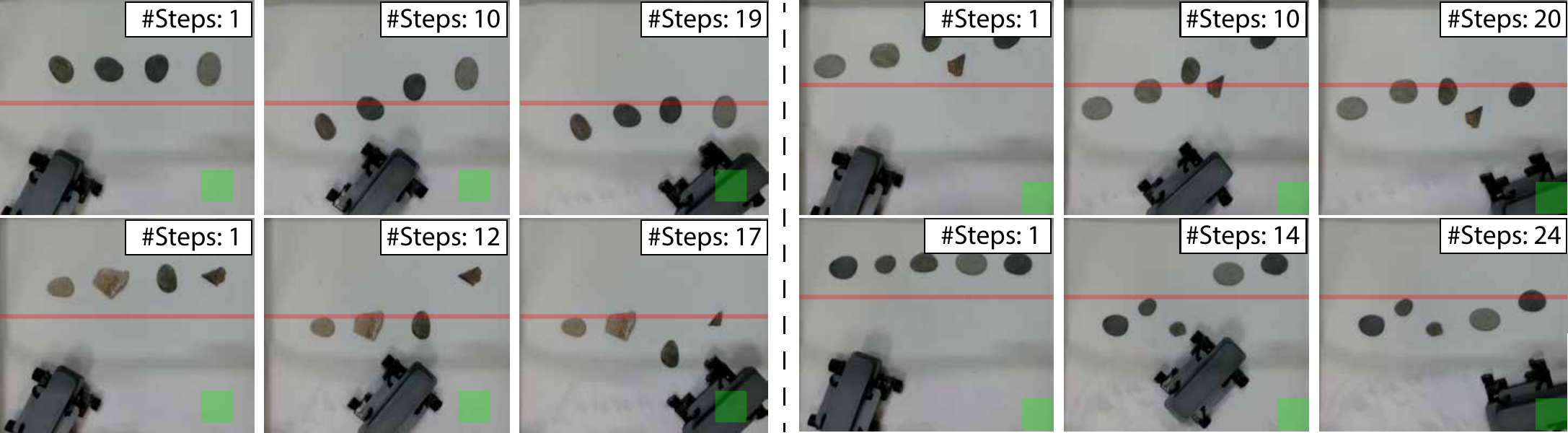}
  \caption{Loco-manipulation (out-of-distribution) experiments with more realistic settings (rocks with different shapes and sizes). We show 4 trials with real rocks, on 16 degrees (top) and 24 degrees (bottom) inclinations.
  }
  \label{fig:experiment real rocks}
  \vspace{-13pt}
\end{figure}

\begin{wraptable}{r}{0.39\textwidth}
    \vspace{-12pt}
    \centering
    \footnotesize
    \begin{tabular}{ccc}
        \toprule
        \textbf{Inclination} & \textbf{MAE}$^\dagger$ & \textbf{Success}\\
        (degree) & (cm) & (\%) \\
        \midrule
        16 & 2.61($\pm$ 1.02) & 65\% \\
        20 & 2.39($\pm$ 0.96) & 70\% \\
        24 & 2.57($\pm$ 0.92) & 65\% \\
        \bottomrule
        \end{tabular}
    \caption{\method performance for loco-manipulation (out-of-distribution).
    }
    \label{tab:DiffusiveGRAIN performance on real rocks}
    \vspace{-12pt}
\end{wraptable}
\emph{Loco-manipulation (out-of-distribution):} 
To test generalization of \method, we conduct experiments using \emph{real rocks} with variations in shape and angularity (see Fig.~\ref{fig:experiment real rocks}), without any fine-tuning. We also test 2 inclination angles (16 and 24 degrees) in addition to 20 degrees tested earlier, to evaluate performance on different slope inclination angles. 
We perform 60 experiment trials with 20 trials for each inclination angle. The experiments show high success rates and small prediction errors (Tab.~\ref{tab:DiffusiveGRAIN performance on real rocks}). 

\section{Conclusion} 
\label{sec:conclusion}
\vspace{-7pt}

We introduce \method, which enables a locomoting legged robot to leverage sand avalanche dynamics to reposition rocks on a sand slope. By integrating a diffusion-based environment state predictor and a robot state predictor, we show that \method is able to achieve a significantly high success rate (70\%, as compared to 20\% using the baseline method) for robot loco-manipulation tasks where a change of robot state needs to be jointly considered and planned with obstacle movement. Our method opens a new avenue for robots to actively manipulate their locomotion environments to improve mobility on challenging granular slopes.

\newpage
\section{Limitations} 
\label{sec:limitations}


One key limitation is that \method is vision-focused and does not capture obstacle properties that may influence avalanche behavior, such as mass. In natural environments, rocks may have significantly larger mass than those used in our experiments, and such properties might not be fully apparent from vision alone. Future work should investigate how avalanche dynamics vary with obstacle mass and how to integrate such information into predictive models. Additionally, the current policy execution performs exhaustive evaluation over all possible action sequences. While effective for small action spaces and short horizons, this brute-force approach becomes computationally infeasible as the problem size grows, and future work could explore approaches to mitigate the computational burden. 
\bibliography{example}  

\clearpage
\input{supplementary}

\end{document}

%% file: supplementary.tex
\appendix

We structure the Appendix as follows:
\begin{itemize}[leftmargin=*,noitemsep]
    \item App.~\ref{app:hardware} presents additional hardware details.
    \item App.~\ref{app:additional_experiments} has additional experimental results. 
    \item App.~\ref{app:baseline} presents and evaluates multiple additional baselines.
    \item App.~\ref{app:details_env_robot} has more details of \method, including information about our environment state predictor, robot state predictor, and cost function. 
\end{itemize}

\section{Additional Details of Hardware Design}
\label{app:hardware}

\begin{figure*}[htbp]
    \centering
    \includegraphics[width=1.0\textwidth]{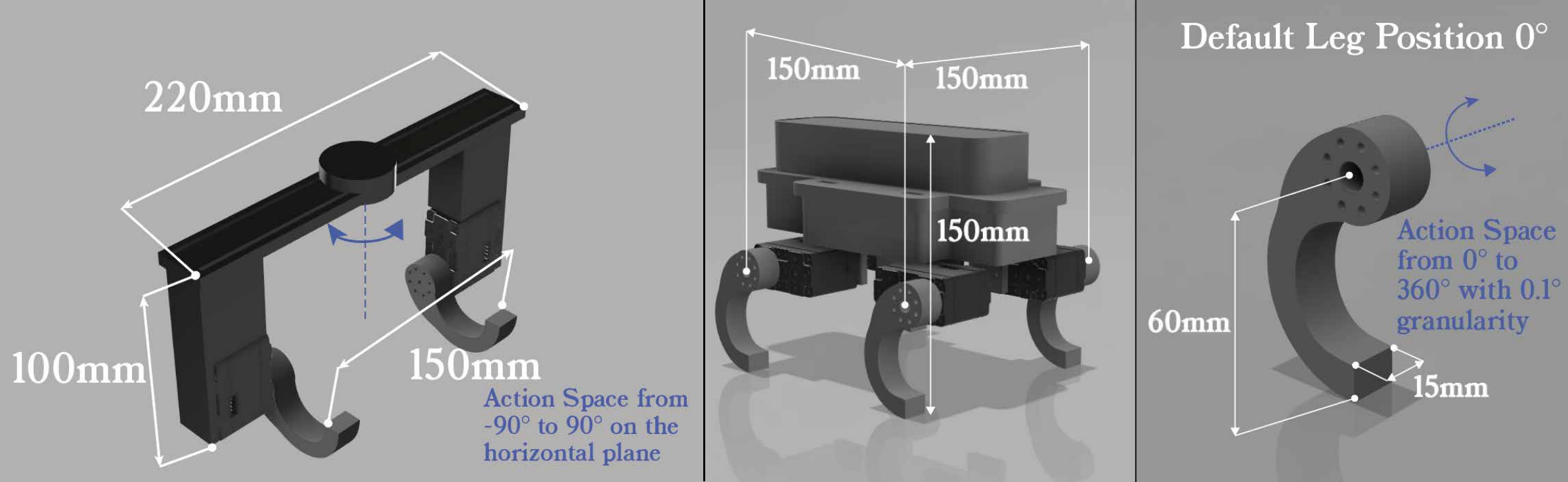}
    \caption{Manipulator and robot design. \emph{Left:} The manipulator is designed to simulate the robot leg action for different robot body sizes. We fixed the distance between the two motors to 150mm in this study to match the robot legs. \emph{Middle :} Sand-proof Robot Design. \emph{Right:} Robot leg design.}
    \label{fig:Slidebar Design}
    \vspace{-5pt}
\end{figure*}

The primary hardware components of this work are the trackway manipulation system and the sand-proof robot. In this design, the manipulator is able to perform horizontal rotations on the xy-plane and lock at specific angles within a range of -90° to 90°. Furthermore, the gantry system allows precise positioning of the manipulators within the sand tank for targeted experimental operations. A custom slide bar was 3D-printed with two slots shown in Fig.~\ref{fig:Slidebar Design}, allowing the motors (Lynxmotion ST) to be mounted to the slide bar and slide within a range of 0–15 cm on either side of the slide bar’s center. For the rotation functionality, we used a magic arm with a lock, attached to the center of the slide bar. The magic arm has two DoFs; we used screws to fix one degree and mounted the arm on the XY table's step motor via a metal bar.

We developed the control unit using an ADA Lynxmotion board and a Raspberry Pi. The unit manages the motion of the XY table and the manipulator's excavation actions. To enhance the flexibility of the experimental procedure, we programmed the Raspberry Pi for remote control via SSH.

The sand-proof robot was designed with compactness as a core principle, prioritizing minimal size while retaining functionality. Key design dimensions are as follows: the distance between the two front legs (or the two rear legs), including the width of the legs, is 15 cm. The robot's height ranges from a maximum of 15 cm (in a fully standing position) to a minimum of 10 cm (when the legs are folded). The maximum length, measured with respect to the longest component, is approximately 15 cm. The robot has a total mass of 849.1 g, including the LiPo battery. The frame of the robot is 3D-printed using PLA material. This compact and efficient design enables the robot to operate effectively within constrained environments. Since the glass beads are extremely fine and pose a risk of damaging the motor mechanisms, protecting the motors from sand intrusion was a critical task. To address this issue, we covered the Lynxmotion Standard Torque motors with tape to ensure good sealing.

\section{Additional Experiments}
\label{app:additional_experiments}

\begin{figure*}[h]
    \centering
    \includegraphics[width=0.8\textwidth]{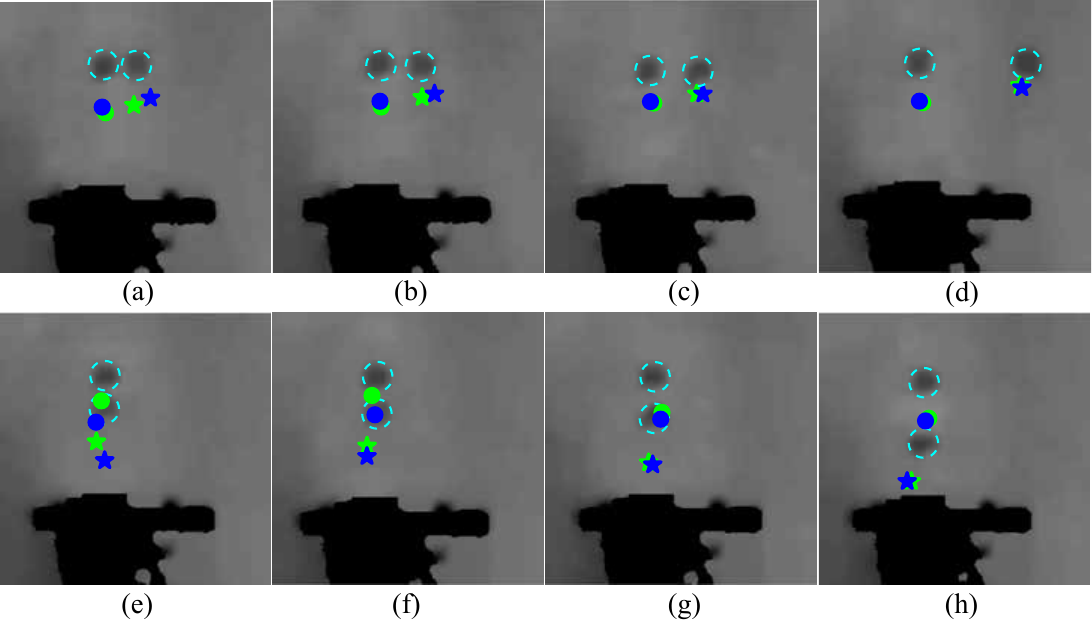}
    \caption{GRAIN performance on obstacles with different lateral or fore-aft distances. (a)(b)(c)(d) are obstacles with a lateral distance of 0cm, 2cm, 4cm, and 8cm respectively; (e)(f)(g)(h) are obstacles with a fore-aft distance of 0cm, 2cm, 4cm, and 8cm respectively. The cyan dashed circles are the obstacle positions before the excavation action, and the green star and circle are the ground truth positions of the obstacles after an excavation action. The blue star and circle are the GRAIN predicted positions of the obstacles after the excavation action. Specifically, the star shape markers are for the reference obstacle and circle shape markers are for the obstacle with different lateral or fore-aft distance to the reference obstacle.}
    \label{fig:example of GRAIN on obstacles with diff distance}
    \vspace{-5pt}
\end{figure*}

\begin{figure*}[htbp]
    \centering
    \includegraphics[width=0.6\textwidth]{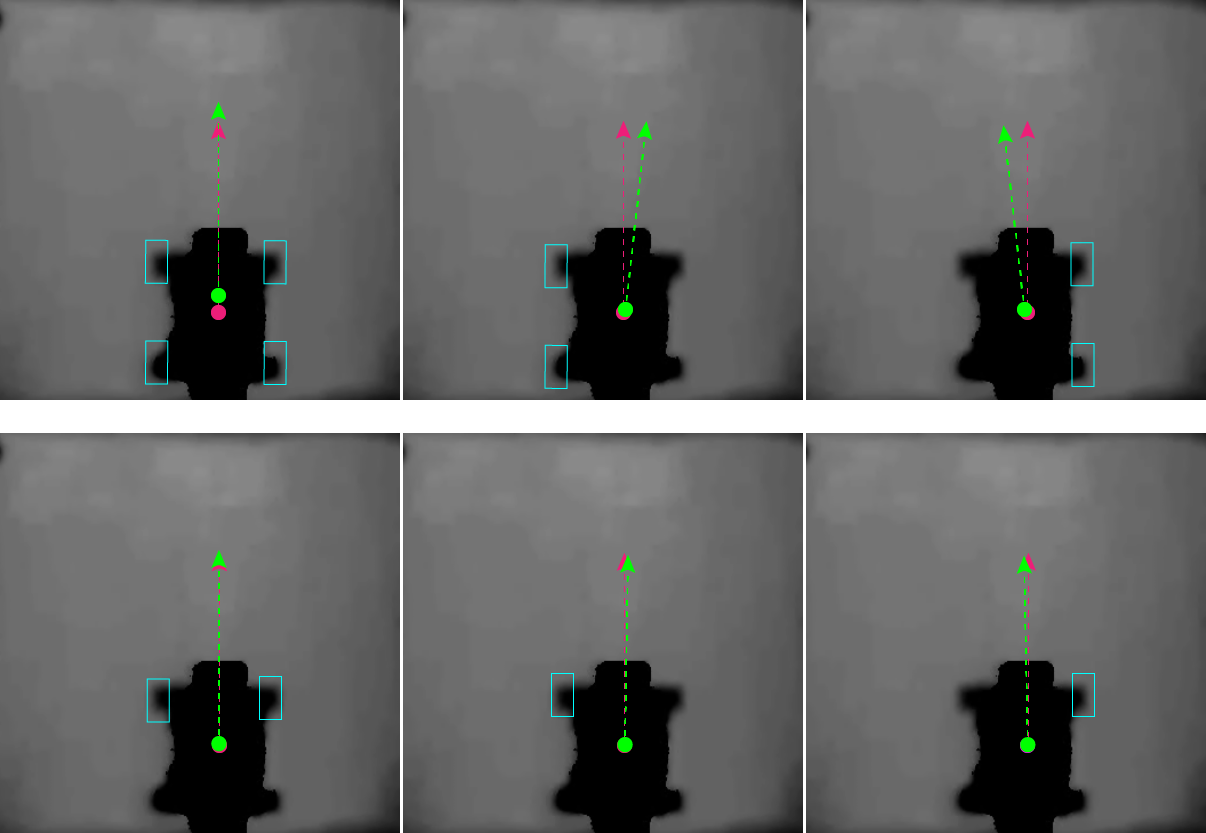}
    \caption{Robot state before and after different robot actions. The cyan rectangle represents the robot action (activated legs), the pink circle with the dashed arrow represents the robot CoM position and orientation before the robot action, and the green circle with the dashed arrow represents the ground truth of the robot CoM position and orientation after the robot action.}
    \label{fig:comparison of GRAIN and current on locomotion}
    \vspace{-5pt}
\end{figure*}

\begin{figure}[h]
    \centering
    \includegraphics[width=0.65\textwidth]{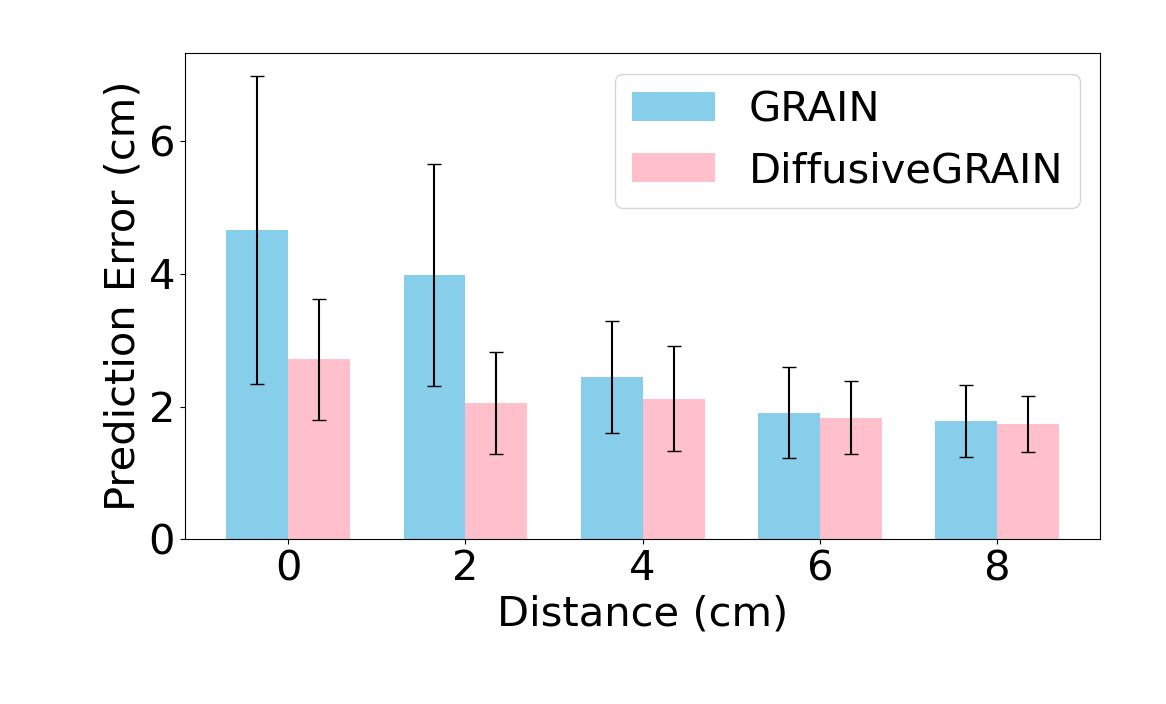}
    \caption{GRAIN prediction errors on obstacles with different distances. The blue and pink bars represent the mean prediction error of GRAIN and \method (our proposed method, see Sec.~\ref{Sec: method}). The error bars indicate the standard deviation.}
    \label{fig:statisitcs of GRAIN on obstacles with diff distance}
    \vspace{-5pt}
\end{figure}

\textbf{Additional experiments for GRAIN}: To test GRAIN's assumptions regarding independent obstacle motion, we evaluate how GRAIN performance changes as a function of the obstacle distance on the sand slope, by calculating its prediction error with different obstacle distances. 
We found that the prediction error worsens when the two obstacles are closer (Fig.~\ref{fig:statisitcs of GRAIN on obstacles with diff distance}). 
Fig.~\ref{fig:example of GRAIN on obstacles with diff distance} illustrates the experiments with varied obstacle distances, and shows that GRAIN's prediction error is low in (Fig.~\ref{fig:example of GRAIN on obstacles with diff distance}, c, d, g, h) as obstacle distances were relatively large, but becomes significantly larger as the distance of obstacles decreases (Fig.~\ref{fig:example of GRAIN on obstacles with diff distance}, a, b, e, f).
Similarly, to test the effect of different manipulation actions on robot state, we performed experiments with different leg manipulation pairs.  Fig.~\ref{fig:comparison of GRAIN and current on locomotion} shows the example of the 6 robot actions where the cyan highlighted legs are activated legs and green arrowed circles represent the robot states after the corresponding action. The experiment indicates we need to consider the robot's state change due to the leg excavation; otherwise, the cumulated robot state changes with multiple steps will introduce errors in predicting robot manipulation results.

\begin{figure*}[htbp]
    \centering
    \includegraphics[width=1.0\textwidth]{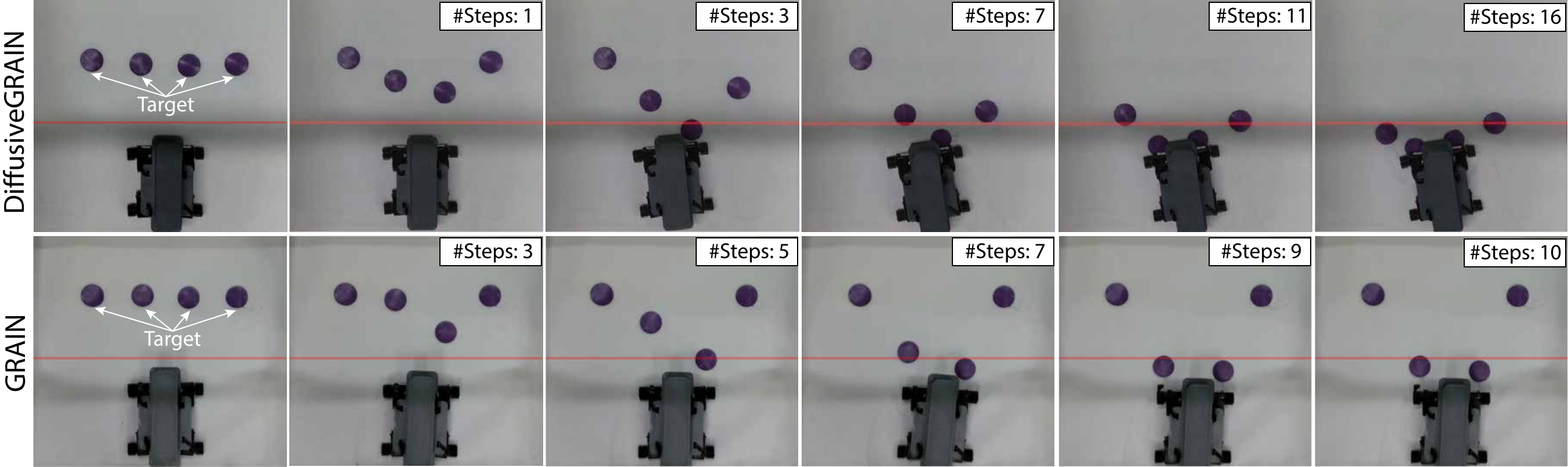}
    \caption{One robot manipulation trial using \method and GRAIN. The robot starts in the bottom middle and must manipulate all 4 obstacles to below the red line. \method results in success, but with GRAIN, the robot only manipulated 2 of 4 obstacles to below the red line, and policy execution terminated as GRAIN predicted no action could further optimize the cost function at Step 10.
    }
    \label{fig:manipulation experiment}
    \vspace{-5pt}
\end{figure*}

\begin{figure*}[htbp]
    \centering
    \includegraphics[width=1.0\textwidth]{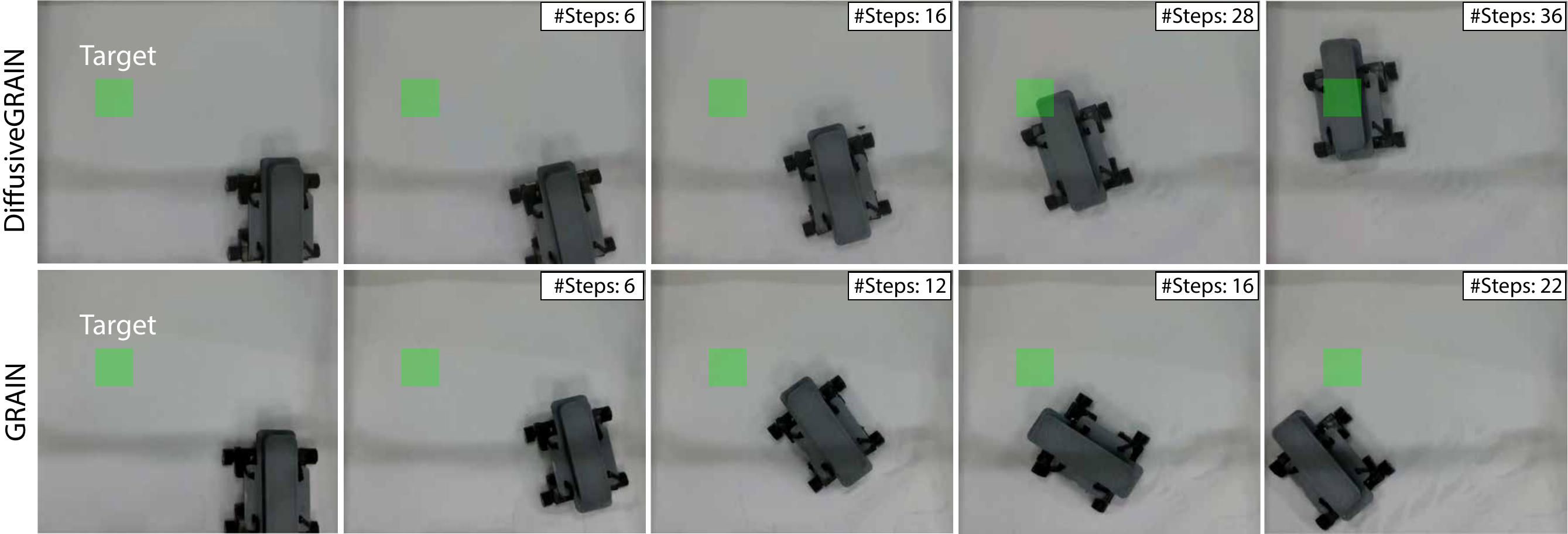}
    \caption{One robot locomotion trial using \method and GRAIN. The robot starts in the bottom right corner and must locomote to the target area marked by a green rectangle. DiffusiveGRAIN results in success, but with GRAIN, the robot reaches the sand tank boundary which is considered a failure.
    }
    \label{fig:locomotion experiment}
    \vspace{-10pt}
\end{figure*}

\textbf{Additional experiments for \method}: We performed experiments to compare the performance between \method and GRAIN. Fig.~\ref{fig:manipulation experiment} shows 1 manipulation trial using \method and GRAIN. In both trials, the robot must (indirectly) manipulate all 4 obstacles to bring them to a designated goal region near the robot. Fig.~\ref{fig:locomotion experiment} shows 1 locomotion trial using \method and GRAIN. We place the robot in the bottom right region, and the robot must move to the area marked by the green rectangle. In both manipulation and locomotion trials, the improved performance suggested the importance for including training data with multiple obstacles, especially when obstacle distances are small (Fig.~\ref{fig:statisitcs of GRAIN on obstacles with diff distance}).

\begin{figure*}[htbp]
    \centering
    \includegraphics[width=1.0\textwidth]{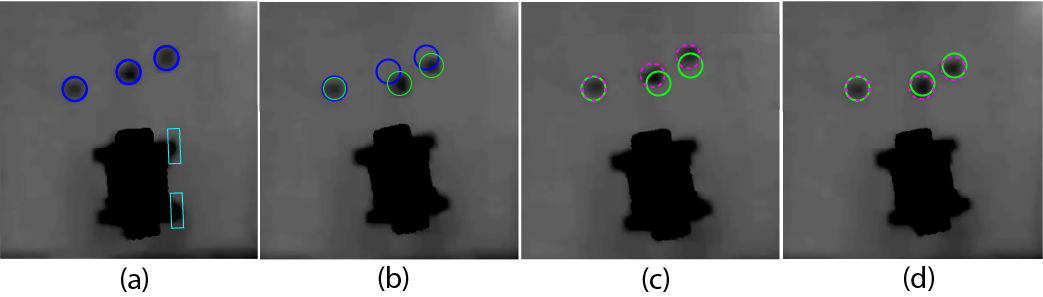}
    \caption{The EAA (see Sec.~\ref{ssec:EAA}). (a) an example of obstacle states before the robot executes a \emph{left turn} action where the blue circles are obstacles and cyan rectangles are the activated legs for the robot; (b) the ground truth of the robot state and obstacle states after the robot executed a \emph{RP} action in (a) where the blue circles and green circles are the obstacle states before and after the robot action; (c) the \emph{Environment Predictor} predicted obstacles states as marked by pink dash circles without EAA; (d) the \emph{Environment Predictor} predicted obstacles states as marked by pink dash circles with EAA.}
    \label{fig:Effective Action Adjustment}
    \vspace{-5pt}
\end{figure*}

\textbf{Additional experiments for EAA}:
Fig.~\ref{fig:Effective Action Adjustment} shows an example of the \method prediction of robot and obstacle states w/o EAA. A statistical evaluation of EAA performance is shown in Tab.~\ref{tab:comparison of performance on dataset for all 5 methods}. 

\section{Additional Baselines}
\label{app:baseline}

\begin{table}[b]
    \centering
    \small
    \begin{tabular}{>{\centering\arraybackslash}m{3cm}m{4.5cm}m{2cm}}
            \toprule
            \textbf{Task} & \textbf{Method} & \textbf{MAE} (cm)\\
            \midrule
            Obstacle state & E2E-Diffusion \newline Pose-Diffusion \newline GRAIN \newline \method \newline \method (without EAA) & 3.81($\pm$ 1.25) \newline 5.04($\pm$ 1.66) \newline 2.91($\pm$ 0.88) \newline 2.44($\pm$ 0.78) \newline 2.80($\pm$ 0.84)\\
            \midrule
            Robot state (position) & E2E-Diffusion \newline Pose-Diffusion \newline GRAIN \newline \method \newline \method (without EAA) & 2.25($\pm$ 0.96) \newline 2.94($\pm$ 1.03) \newline 1.26($\pm$ 0.41)  \newline 1.17($\pm$ 0.32) \newline 1.17($\pm$ 0.32)\\
            \bottomrule
        \end{tabular}
    \vspace{6pt}
    \caption{
        We compare the performance of the E2E-Diffusion, Pose-Diffusion, GRAIN, \method, and \method (without EAA) on the experiment dataset.
    }
    \label{tab:comparison of performance on dataset for all 5 methods}
    \vspace{-12pt}
\end{table}

\emph{E2E-Diffusion}: In this paper, we use two models: an environment state predictor and a robot state predictor. To investigate if this division is necessary, we design another baseline (E2E-Diffusion) which is a single diffusion model for end-to-end prediction. Specifically, we retrained the same U-Net-based diffusion model in the \emph{Environment State Predictor} on all 300 trials in our data, and have it output the predicted depth image at the next state. We evaluated E2E-Diffusion on the experiment dataset by extracting the predicted obstacle and robot states from the predicted depth image at the next state and comparing them with the ground truth. 

\emph{Pose-Diffusion}: To test our proposed representations of the system, we trained a new version of the environment state predictor and robot state predictor on pose inputs, which are solely the robot and obstacle states. Like \method, this Pose-Diffusion baseline also outputs the predicted robot and obstacle states. The input to the network is a $B * (3 + 2 * N + 1$) vector, and the output is a vector with the dimension of $B * (3 + 2 * N$), where the input vector is a concatenation of the robot state, a 1 by 3 vector, and $N$ obstacle states, a 1 by $2N$ vector, and the indicator of robot action, a 1 by 1 vector (an integer from 0 to 5 represents the 6 actions in this study). In addition, $B$ is the batch size; we used $B = 16$.

\emph{\method (without EAA)}: As an ablation, we also test the \method without EAA in the policy execution for the experiment data. 

We report the performance of E2E-Diffusion, Pose-Diffusion, GRAIN, \method, and \method (without EAA) in Tab.~\ref{tab:comparison of performance on dataset for all 5 methods}. As shown in Tab.~\ref{tab:comparison of performance on dataset for all 5 methods}, \method outperforms \method (without EAA), E2E-Diffusion, and GRAIN on the experiment dataset. We hypothesize that the low performance of E2E-Diffusion is because it must learn from a highly complex dataset from combining the manipulator and robot datasets. Consequently, this leads to the current dataset size being insufficient to achieve the same performance level as compared to \method (which may be more data-efficient). We plan to investigate this hypothesis in the future by collecting more training data in experiments. 

The performance of Pose-Diffusion implies that solely relying on the pose representation does not work well. Our experiments showed that an obstacle with the same initial pose could have significantly different displacements in response to the same manipulation action, \emph{depending on the terrain state}. This aligned with previous literature~\cite{daerr1999two} indicating that the terrain surface significantly affects obstacle displacement with the same manipulation action.

\begin{figure*}[h]
    \centering
    \includegraphics[width=1.0\textwidth]{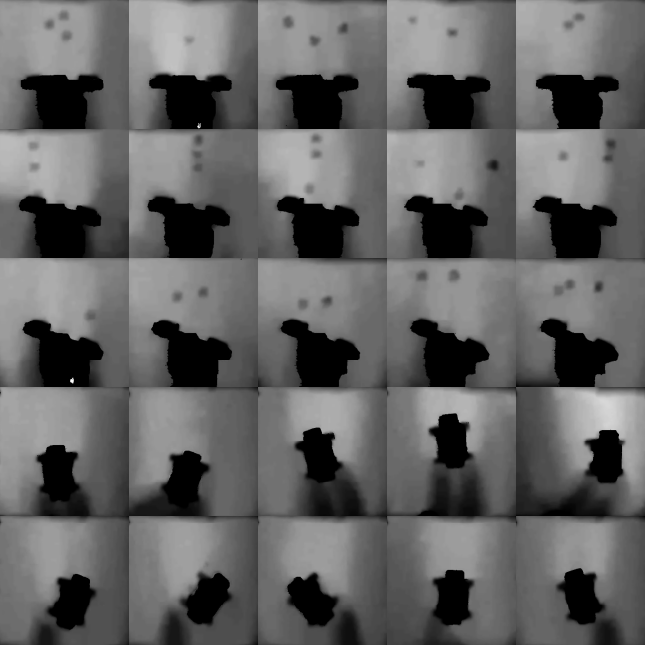}
    \caption{Samples of the dataset.}
    \label{fig:dataset samples}
    \vspace{-5pt}
\end{figure*}

\section{Additional Details of \method}
\label{app:details_env_robot}

In \method, we trained the environment state predictor $f_e$ using the dataset collected by the manipulator. During policy execution, $f_e$ takes depth image inputs from the robot. Although the manipulator and the robot differ in size and appear slightly different in the depth images, they share common characteristics. We believe the diffusion model learns to ignore these appearance differences and focuses on predicting changes in the environment state.

The robot state predictor $f_r$ is trained on the dataset collected by the robot. To make the network robust to inputs during policy execution, we augmented the training data by adding obstacles into the depth images. The primarily consideration to decouple $f_e$ and $f_r$ is that it significantly reduces the complexity of the loco-manipulation problem, which in turn lowers the amount of data needed to train the networks, which is demonstrated in Tab.~\ref{tab:comparison of performance on dataset for all 5 methods}. This is critical in our study, as all data were collected from real physical experiments.
Our preliminary experiments suggested that obstacle positions on the granular slope have little effect on the robot’s state change under a given action when the distance between robot leg and obstacle is relatively large, which supports the chosen method of obstacle augmentation. We acknowledge that in the case of direct leg-obstacle contact, the current method may not accurately predict robot and obstacle state changes, which is a limitation of this study that future work should further explore. 

\begin{figure*}[h]
    \centering
    \includegraphics[width=1.00\textwidth]{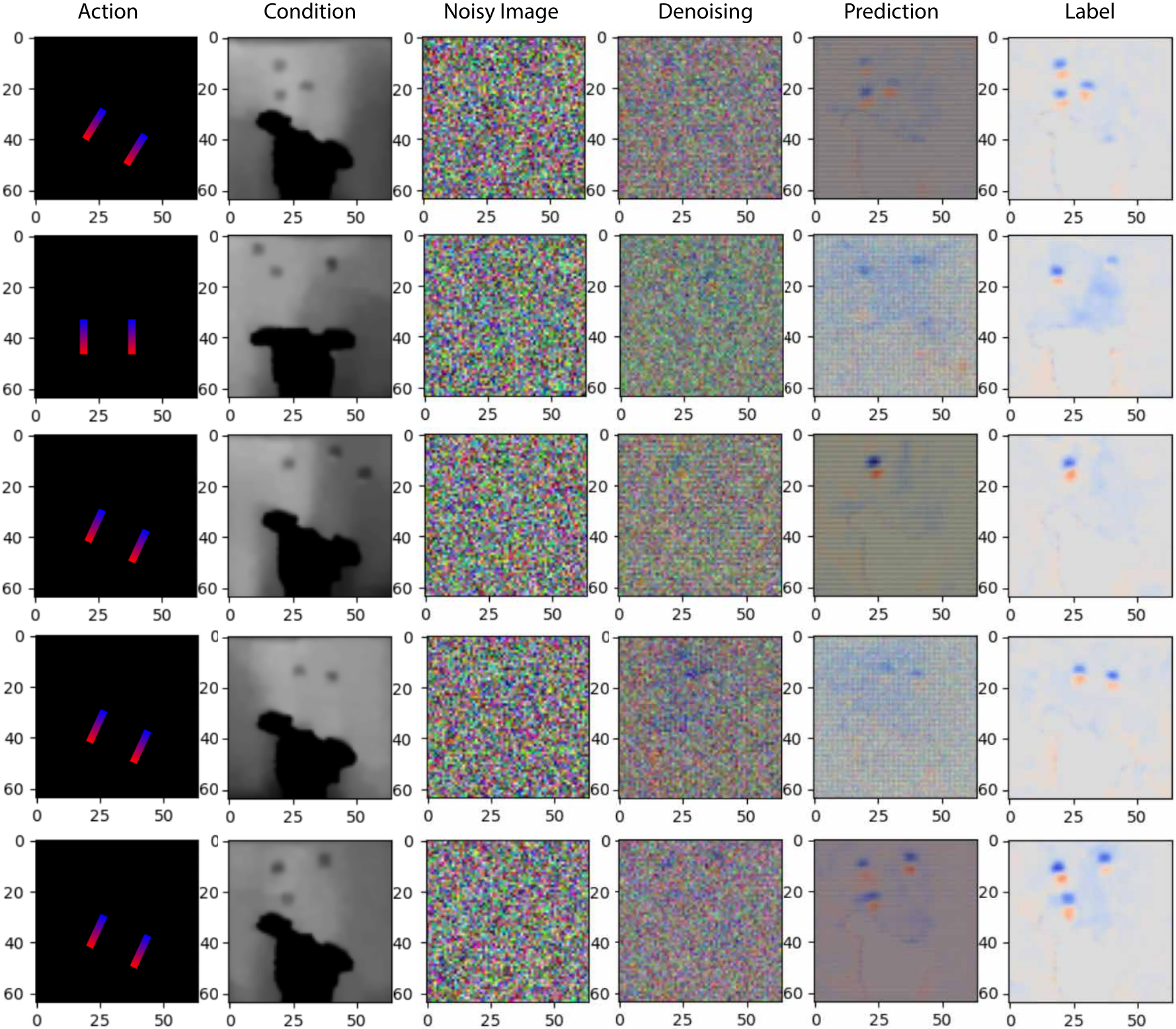}
    \caption{Examples of \emph{Environment Predictor} prediction (one per row). From left to right are the action image $\bA_t$, input image $\bI_t$, pure noise image, the denoised image at diffusion step 100, output image, and label image respectively.}
    \label{fig:manipulation model predictions}
    \vspace{-5pt}
\end{figure*}

\begin{figure*}[h]
    \centering
    \includegraphics[width=0.75\textwidth]{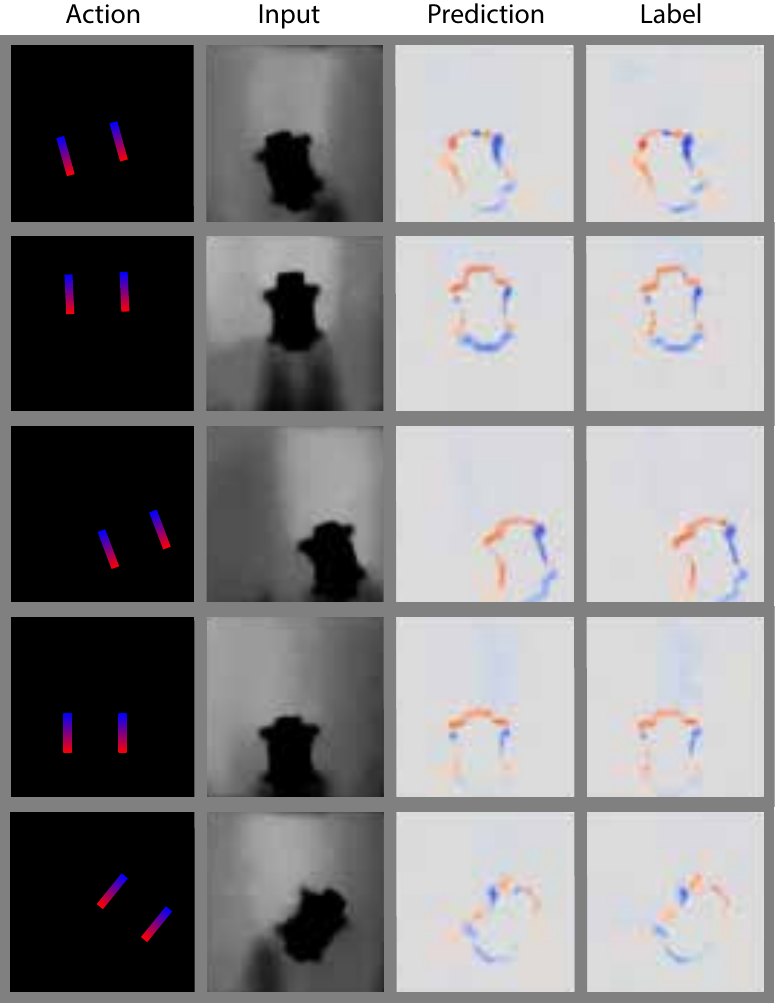}
    \caption{Examples of \emph{Robot State Predictor} prediction (one per row). From the left to right are the action image $\bA_t$, input image $\bI_t$, output image, and label image respectively.}
    \label{fig:locomotion model predictions}
    \vspace{-5pt}
\end{figure*}

We show some examples of the training dataset in Fig.~\ref{fig:dataset samples}, and we also show some output images of the environment state predictor and robot state predictor on the validation dataset in Fig.~\ref{fig:manipulation model predictions} and Fig.~\ref{fig:locomotion model predictions}, respectively. We report the key hyperparameters of $f_e$ and $f_r$ in Tab.~\ref{tab:diffusion-hyperparams} and Tab.~\ref{tab:unet-hyperparams}, respectively.

In addition, we provide additional details on the cost function used for loco-manipulation (Sec. \ref{sec:costfun}): For a robot that does both locomotion and manipulation, we linearly combine the cost of locomotion and manipulation to get the cost for the robot ``loco-manipulation'' task: $\bC_{lo} = w_5 \bC_l + w_6 \bC_m$, where in this study, we use $w_5 = 0.4$ and $w_6 = 0.6$. 

\begin{table}[h]
    \centering
    \small
    \begin{tabular}{>{\centering\arraybackslash}m{4cm}m{4cm}m{4cm}}
        \toprule
        \textbf{Category} & \textbf{Hyperparameter} & \textbf{Value}\\
        \midrule
        \multirow{3}{*}{\shortstack[l]{Diffusion Process}}
        & \# of Diffusion Steps & 150 \\
        & Beta Schedule & Linear (0.0001 $\to$ 0.025) \\
        & Sampling Algorithm & DDPM \\
        \midrule
        \multirow{4}{*}{\shortstack[l]{U-Net Architecture}}
        & Base Channels & 64  \\
        & Channel Multiplier & [1, 2, 4, 8, 16]  \\
        & \# of Res. Blocks & 2  \\
        & Dropout & 0.1  \\
        \midrule
        \multirow{2}{*}{Time Embedding}
        & Embedding Dim & 32  \\
        & Activation & ReLU  \\
        \midrule
        \multirow{5}{*}{\shortstack[l]{Optimizer and Training}}
        & Optimizer & AdamW \\
        & Learning Rate & 1$\times 10^{-5}$  \\
        & Weight Decay & 1$\times 10^{-5}$  \\
        & Batch Size & 1  \\
        \midrule
        \multirow{2}{*}{Data}
        & Resolution & 400$\times$400  \\
        & Normalization & $[-1, 1]$  \\
        \bottomrule
    \end{tabular}
    \vspace{5pt}
    \caption{
        Hyperparameters of the Environment State Predictor $f_e$. 
    }
    \label{tab:diffusion-hyperparams}
    \vspace{-10pt}
\end{table}

\begin{table}[h]
    \centering
    \small
    \begin{tabular}{>{\centering\arraybackslash}m{4cm}m{4cm}m{3cm}}
        \toprule
        \textbf{Category} & \textbf{Hyperparameter} & \textbf{Value}\\
        \midrule
        \multirow{4}{*}{\shortstack[l]{U-Net Architecture}}
        & Base Channels & 64  \\
        & Channel Multiplier & [1, 2, 4, 8]  \\
        & \# of Res. Blocks & 2  \\
        & Activation & ReLU  \\
        & Dropout & 0.1  \\
        \midrule
        \multirow{5}{*}{\shortstack[l]{Optimizer and Training}}
        & Optimizer & AdamW \\
        & Learning Rate & 1$\times 10^{-4}$  \\
        & Weight Decay & 1$\times 10^{-4}$  \\
        & Batch Size & 1  \\
        \midrule
        \multirow{2}{*}{Data}
        & Resolution & 400$\times$400  \\
        & Normalization & $[-1, 1]$  \\
        \bottomrule
    \end{tabular}
    \vspace{5pt}
    \caption{
        Hyperparameters of the Robot State Predictor $f_r$. 
    }
    \label{tab:unet-hyperparams}
    \vspace{-10pt}
\end{table}

%% file: main.bbl
\begin{thebibliography}{36}
\providecommand{\natexlab}[1]{#1}
\providecommand{\url}[1]{\texttt{#1}}
\expandafter\ifx\csname urlstyle\endcsname\relax
  \providecommand{\doi}[1]{doi: #1}\else
  \providecommand{\doi}{doi: \begingroup \urlstyle{rm}\Url}\fi

\bibitem[Schiebel et~al.(2017)Schiebel, Rieser, Hubbard, Chen, and Goldman]{schiebel2017collisional}
P.~E. Schiebel, J.~M. Rieser, A.~M. Hubbard, L.~Chen, and D.~I. Goldman.
\newblock Collisional diffraction emerges from simple control of limbless locomotion.
\newblock In \emph{Conference on Biomimetic and Biohybrid Systems}, pages 611--618. Springer, 2017.

\bibitem[Schiebel et~al.(2019)Schiebel, Rieser, Hubbard, Chen, Rocklin, and Goldman]{schiebel2019mechanical}
P.~E. Schiebel, J.~M. Rieser, A.~M. Hubbard, L.~Chen, D.~Z. Rocklin, and D.~I. Goldman.
\newblock Mechanical diffraction reveals the role of passive dynamics in a slithering snake.
\newblock \emph{Proceedings of the National Academy of Sciences}, 116\penalty0 (11):\penalty0 4798--4803, 2019.

\bibitem[Qian and Koditschek(2020)]{qian2019obstacle}
F.~Qian and D.~E. Koditschek.
\newblock An obstacle disturbance selection framework: emergent robot steady states under repeated collisions.
\newblock \emph{The International Journal of Robotics Research}, 2020.

\bibitem[Othayoth et~al.(2020)Othayoth, Thoms, and Li]{othayoth2020energy}
R.~Othayoth, G.~Thoms, and C.~Li.
\newblock An energy landscape approach to locomotor transitions in complex 3d terrain.
\newblock \emph{Proceedings of the National Academy of Sciences}, 117\penalty0 (26):\penalty0 14987--14995, 2020.

\bibitem[Rieser et~al.(2019)Rieser, Schiebel, Pazouki, Qian, Goddard, Wiesenfeld, Zangwill, Negrut, and Goldman]{rieser2019}
J.~M. Rieser, P.~E. Schiebel, A.~Pazouki, F.~Qian, Z.~Goddard, K.~Wiesenfeld, A.~Zangwill, D.~Negrut, and D.~I. Goldman.
\newblock Dynamics of scattering in undulatory active collisions.
\newblock \emph{Physical Review E}, 99\penalty0 (2):\penalty0 022606, 2019.

\bibitem[Wang et~al.(2023)Wang, Pierce, Kojouharov, Chong, Diaz, Lu, and Goldman]{wang2023mechanical}
T.~Wang, C.~Pierce, V.~Kojouharov, B.~Chong, K.~Diaz, H.~Lu, and D.~I. Goldman.
\newblock Mechanical intelligence simplifies control in terrestrial limbless locomotion.
\newblock \emph{arXiv preprint arXiv:2304.08652}, 2023.

\bibitem[Hu and Qian(2024)]{haodi-obstacle}
H.~Hu and F.~Qian.
\newblock Obstacle-aided trajectory control of a quadrupedal robot through sequential gait composition.
\newblock \emph{IEEE Transactions on Robotics}, pages 1--15, 2024.

\bibitem[Hu et~al.(2024)Hu, Liao, Du, and Qian]{hu2024multirobot}
H.~Hu, X.~Liao, W.~Du, and F.~Qian.
\newblock Multi-robot connection towards collective obstacle field traversal, 2024.
\newblock URL \url{https://arxiv.org/abs/2409.11709}.

\bibitem[Ramesh et~al.(2020)Ramesh, Kathail, Koditschek, and Qian]{ramesh2020modulation}
D.~Ramesh, A.~Kathail, D.~E. Koditschek, and F.~Qian.
\newblock Modulation of robot orientation via leg-obstacle contact positions.
\newblock \emph{IEEE Robotics and Automation Letters}, 5\penalty0 (2):\penalty0 2054--2061, 2020.

\bibitem[Chakraborty et~al.(2022)Chakraborty, Hu, Kvalheim, and Qian]{chakraborty2022planning}
K.~Chakraborty, H.~Hu, M.~D. Kvalheim, and F.~Qian.
\newblock Planning of obstacle-aided navigation for multi-legged robots using a sampling-based method over directed graphs.
\newblock \emph{IEEE Robotics and Automation Letters}, 7\penalty0 (4):\penalty0 8861--8868, 2022.

\bibitem[Qian and Goldman(2015)]{qian2015anticipatory}
F.~Qian and D.~Goldman.
\newblock Anticipatory control using substrate manipulation enables trajectory control of legged locomotion on heterogeneous granular media.
\newblock In \emph{Micro-and Nanotechnology Sensors, Systems, and Applications VII}, volume 9467, page 94671U. International Society for Optics and Photonics, 2015.

\bibitem[Hu et~al.(2024)Hu, Qian, and Seita]{hulearning}
H.~Hu, F.~Qian, and D.~Seita.
\newblock Learning granular media avalanche behavior for indirectly manipulating obstacles on a granular slope.
\newblock In \emph{8th Annual Conference on Robot Learning}, 2024.

\bibitem[Barker and Mehta(2000)]{barker2000two}
G.~Barker and A.~Mehta.
\newblock Two types of avalanche behaviour in model granular media.
\newblock \emph{Physica A: Statistical Mechanics and its Applications}, 283\penalty0 (3-4):\penalty0 328--336, 2000.

\bibitem[Pudasaini and Hutter(2007)]{pudasaini2007avalanche}
S.~P. Pudasaini and K.~Hutter.
\newblock \emph{Avalanche dynamics: dynamics of rapid flows of dense granular avalanches}.
\newblock Springer Science \& Business Media, 2007.

\bibitem[Gravish and Goldman(2014)]{gravish2014effect}
N.~Gravish and D.~I. Goldman.
\newblock Effect of volume fraction on granular avalanche dynamics.
\newblock \emph{Physical Review E}, 90\penalty0 (3):\penalty0 032202, 2014.

\bibitem[Pavlov and Johnson(2019)]{pavlov2019soil}
C.~Pavlov and A.~M. Johnson.
\newblock Soil displacement terramechanics for wheel-based trenching with a planetary rover.
\newblock In \emph{2019 International Conference on Robotics and Automation (ICRA)}, pages 4760--4766. IEEE, 2019.

\bibitem[Wang et~al.(2023)Wang, Li, Driggs-Campbell, Fei-Fei, and Wu]{wang2023dynamicresolution}
Y.~Wang, Y.~Li, K.~Driggs-Campbell, L.~Fei-Fei, and J.~Wu.
\newblock {Dynamic-Resolution Model Learning for Object Pile Manipulation}.
\newblock In \emph{Robotics: Science and Systems (RSS)}, 2023.

\bibitem[Xue et~al.(2023)Xue, Cheng, Kachana, and Xu]{xue2023neuralfield}
S.~Xue, S.~Cheng, P.~Kachana, and D.~Xu.
\newblock {Neural Field Dynamics Model for Granular Object Piles Manipulation}.
\newblock In \emph{Conference on Robot Learning (CoRL)}, 2023.

\bibitem[Schenck et~al.(2017)Schenck, Tompson, Levine, and Fox]{schenck2017learning}
C.~Schenck, J.~Tompson, S.~Levine, and D.~Fox.
\newblock Learning robotic manipulation of granular media.
\newblock In \emph{Conference on Robot Learning (CoRL)}, 2017.

\bibitem[Ronneberger et~al.(2015)Ronneberger, Fischer, and Brox]{ronneberger2015unet}
O.~Ronneberger, P.~Fischer, and T.~Brox.
\newblock {U-Net: Convolutional Networks for Biomedical Image Segmentation}.
\newblock In \emph{International Conference on Medical Image Computing and Computer-Assisted Intervention (MICCAI)}, 2015.

\bibitem[Sohl-Dickstein et~al.(2015)Sohl-Dickstein, Weiss, Maheswaranathan, and Ganguli]{sohldickstein2015deepunsupervisedlearningusing}
J.~Sohl-Dickstein, E.~A. Weiss, N.~Maheswaranathan, and S.~Ganguli.
\newblock Deep unsupervised learning using nonequilibrium thermodynamics.
\newblock In \emph{International Conference on Machine Learning (ICML)}, 2015.

\bibitem[Yang et~al.(2023)Yang, Zhang, Song, Hong, Xu, Zhao, Zhang, Cui, and Yang]{yang2023diffusion}
L.~Yang, Z.~Zhang, Y.~Song, S.~Hong, R.~Xu, Y.~Zhao, W.~Zhang, B.~Cui, and M.-H. Yang.
\newblock Diffusion models: A comprehensive survey of methods and applications.
\newblock \emph{ACM Computing Surveys}, 56\penalty0 (4):\penalty0 1--39, 2023.

\bibitem[Ho et~al.(2020)Ho, Jain, and Abbeel]{ho2020denoisingdiffusionprobabilisticmodels}
J.~Ho, A.~Jain, and P.~Abbeel.
\newblock {Denoising Diffusion Probabilistic Models}.
\newblock In \emph{Neural Information Processing Systems (NeurIPS)}, 2020.

\bibitem[Maladen et~al.(2009)Maladen, Ding, Li, and Goldman]{maladen2009undulatory}
R.~D. Maladen, Y.~Ding, C.~Li, and D.~I. Goldman.
\newblock Undulatory swimming in sand: subsurface locomotion of the sandfish lizard.
\newblock \emph{science}, 325\penalty0 (5938):\penalty0 314--318, 2009.

\bibitem[Li et~al.(2013)Li, Zhang, and Goldman]{li2013terradynamics}
C.~Li, T.~Zhang, and D.~I. Goldman.
\newblock A terradynamics of legged locomotion on granular media.
\newblock \emph{science}, 339\penalty0 (6126):\penalty0 1408--1412, 2013.

\bibitem[Finn et~al.(2016)Finn, Li, and Apte]{finn2016particle}
J.~R. Finn, M.~Li, and S.~V. Apte.
\newblock Particle based modelling and simulation of natural sand dynamics in the wave bottom boundary layer.
\newblock \emph{Journal of Fluid Mechanics}, 796:\penalty0 340--385, 2016.

\bibitem[Albert et~al.(1997)Albert, Albert, Hornbaker, Schiffer, and Barab{\'a}si]{albert1997maximum}
R.~Albert, I.~Albert, D.~Hornbaker, P.~Schiffer, and A.-L. Barab{\'a}si.
\newblock Maximum angle of stability in wet and dry spherical granular media.
\newblock \emph{Physical Review E}, 56\penalty0 (6):\penalty0 R6271, 1997.

\bibitem[Shrivastava et~al.(2020)Shrivastava, Karsai, Aydin, Pettinger, Bluethmann, Ambrose, and Goldman]{shrivastava2020material}
S.~Shrivastava, A.~Karsai, Y.~O. Aydin, R.~Pettinger, W.~Bluethmann, R.~O. Ambrose, and D.~I. Goldman.
\newblock Material remodeling and unconventional gaits facilitate locomotion of a robophysical rover over granular terrain.
\newblock \emph{Science robotics}, 5\penalty0 (42):\penalty0 eaba3499, 2020.

\bibitem[Karsai et~al.(2022)Karsai, Kerimoglu, Soto, Ha, Zhang, and Goldman]{karsai2022real}
A.~Karsai, D.~Kerimoglu, D.~Soto, S.~Ha, T.~Zhang, and D.~I. Goldman.
\newblock Real-time remodeling of granular terrain for robot locomotion.
\newblock \emph{Advanced Intelligent Systems}, 4\penalty0 (12):\penalty0 2200119, 2022.

\bibitem[Kerimoglu et~al.(2024)Kerimoglu, Soto, Hemsley, Brunner, Ha, Zhang, and Goldman]{kerimoglu2024learning}
D.~Kerimoglu, D.~Soto, M.~L. Hemsley, J.~Brunner, S.~Ha, T.~Zhang, and D.~I. Goldman.
\newblock Learning manipulation of steep granular slopes for fast mini rover turning.
\newblock In \emph{2024 IEEE International Conference on Robotics and Automation (ICRA)}, pages 16985--16990. IEEE, 2024.

\bibitem[Schwarz et~al.(2016)Schwarz, Beul, Droeschel, Sch{\"u}ller, Periyasamy, Lenz, Schreiber, and Behnke]{schwarz2016supervised}
M.~Schwarz, M.~Beul, D.~Droeschel, S.~Sch{\"u}ller, A.~S. Periyasamy, C.~Lenz, M.~Schreiber, and S.~Behnke.
\newblock Supervised autonomy for exploration and mobile manipulation in rough terrain with a centaur-like robot.
\newblock \emph{Frontiers in Robotics and AI}, 3:\penalty0 57, 2016.

\bibitem[Dosovitskiy et~al.(2021)Dosovitskiy, Beyer, Kolesnikov, Weissenborn, Zhai, Unterthiner, Dehghani, Minderer, Heigold, Gelly, et~al.]{dosovitskiy2020image}
A.~Dosovitskiy, L.~Beyer, A.~Kolesnikov, D.~Weissenborn, X.~Zhai, T.~Unterthiner, M.~Dehghani, M.~Minderer, G.~Heigold, S.~Gelly, et~al.
\newblock An image is worth 16x16 words: Transformers for image recognition at scale.
\newblock In \emph{International Conference on Learning Representations (ICLR)}, 2021.

\bibitem[B{\"o}rzs{\"o}nyi et~al.(2008)B{\"o}rzs{\"o}nyi, Halsey, and Ecke]{borzsonyi2008avalanche}
T.~B{\"o}rzs{\"o}nyi, T.~C. Halsey, and R.~E. Ecke.
\newblock Avalanche dynamics on a rough inclined plane.
\newblock \emph{Physical Review E—Statistical, Nonlinear, and Soft Matter Physics}, 78\penalty0 (1):\penalty0 011306, 2008.

\bibitem[Bradski(2000)]{opencv_library}
G.~Bradski.
\newblock {The OpenCV Library}.
\newblock \emph{Dr. Dobb's Journal of Software Tools}, 2000.

\bibitem[Qian and Goldman(2015)]{qian2015dynamics}
F.~Qian and D.~I. Goldman.
\newblock The dynamics of legged locomotion in heterogeneous terrain: universality in scattering and sensitivity to initial conditions.
\newblock In \emph{Robotics: Science and Systems (RSS)}, 2015.

\bibitem[Daerr and Douady(1999)]{daerr1999two}
A.~Daerr and S.~Douady.
\newblock Two types of avalanche behaviour in granular media.
\newblock \emph{Nature}, 399\penalty0 (6733):\penalty0 241--243, 1999.

\end{thebibliography}
